\documentclass{article}

\usepackage{amssymb}
\usepackage{graphicx}

 \usepackage[eandd, final, nonatbib]{neurips_2026}

\usepackage[utf8]{inputenc} 
\usepackage[T1]{fontenc}    
\PassOptionsToPackage{hyphens}{url} 
\usepackage{hyperref}       
\usepackage{url}            
\usepackage{booktabs}       
\usepackage{amsmath}        
\usepackage{amsfonts}       
\usepackage{nicefrac}       
\usepackage{microtype}      
\usepackage{xcolor}         
\usepackage{subcaption}     
\usepackage{longtable}      
\usepackage{xltabular}      
\usepackage{wrapfig}        
\usepackage{makecell}       
\usepackage{colortbl}       
\usepackage{placeins}       

\definecolor{TableHeader}{RGB}{232,239,248}
\definecolor{TableSubHeader}{RGB}{244,247,251}
\definecolor{TableGroup}{RGB}{238,241,245}
\definecolor{TableStripe}{RGB}{249,250,252}
\definecolor{MedEmbRow}{RGB}{255,248,241}
\definecolor{BestCell}{RGB}{255,236,214}
\definecolor{MetricGray}{RGB}{92,92,92}

\title{MedHEB: Benchmarking Medical Embeddings Across Heterogeneous Clinical Evidence}

\author{%
  \makebox[0.30\textwidth][c]{Yingshu Li\thanks{These authors contributed equally to this work.}} \\
  The University of Sydney\\
  Sydney, Australia\\
  {\footnotesize\texttt{yingshu.li@sydney.edu.au}} \\
  \And
  \makebox[0.30\textwidth][c]{Shaoyang Zhou\footnotemark[1]} \\
  The University of Sydney\\
  Sydney, Australia\\
  {\footnotesize\texttt{szho0189@uni.sydney.edu.au}} \\
  \And
  \makebox[0.30\textwidth][c]{Zhanyu Wang} \\
  The University of Sydney\\
  Sydney, Australia\\
  {\footnotesize\texttt{zhanyu.wang@sydney.edu.au}} \\
  \AND
  \makebox[0.30\textwidth][c]{Yunyi Liu} \\
  The University of Sydney\\
  Sydney, Australia\\
  {\footnotesize\texttt{yunyi.liu1@sydney.edu.au}} \\
  \And
  \makebox[0.30\textwidth][c]{Xinyu Liang} \\
  College of Medical Imaging\\
  Shandong Medical and\\
  Pharmaceutical University\\
  Yantai, China\\
  {\footnotesize\texttt{xinyu.liang@sdmpu.edu.cn}} \\
  \And
  \makebox[0.30\textwidth][c]{Xi Zhang} \\
  University of Glasgow\\
  Glasgow, UK\\
  {\footnotesize\texttt{x.zhang.6@research.gla.ac.uk}} \\
  \AND
  \makebox[0.30\textwidth][c]{Lingqiao Liu} \\
  Adelaide University\\
  Adelaide, Australia\\
  {\footnotesize\texttt{lingqiao.liu@adelaide.edu.au}} \\
  \And
  \makebox[0.30\textwidth][c]{Lei Wang} \\
  University of Wollongong\\
  Wollongong, Australia\\
  {\footnotesize\texttt{leiw@uow.edu.au}} \\
  \And
  \makebox[0.30\textwidth][c]{Luping Zhou\thanks{Corresponding author.}} \\
  The University of Sydney\\
  Sydney, Australia\\
  {\footnotesize\texttt{luping.zhou@sydney.edu.au}} \\
}

\begin{document}

\maketitle

\begin{abstract}
Clinical evidence is inherently relational: medical images are interpreted with prior studies, cross-modal examinations, reports, and localized findings, and retrieving relevant evidence is therefore central to clinical interpretation. Recent foundation models increasingly encode these heterogeneous sources into shared embedding spaces, yet existing embedding evaluation often focuses on isolated matches, leaving unclear whether current models retrieve relevant evidence across changes in modality, appearance, and granularity.
We introduce MedHEB, a unified benchmark with $107$ retrieval tasks from $54$ datasets, including both in-distribution and held-out-source evaluation, across three axes: evidence form (2D images, 3D volumes, text), pairing type (image--image, image--text, text--text, and modality-bridging visual matching), and relevance granularity (global, question-conditioned, region-level). 
Under a single ranking protocol, MedHEB varies clinically relevant evidence definitions to test whether embedding proximity remains meaningful across heterogeneous evidence, including modality-bridging visual retrieval and region-level text-to-region retrieval. 
Across representative models, performance is uneven across retrieval families: models strong on conventional image--text or text--text retrieval are less reliable on modality-bridging and region-level tasks. 
Current embedding spaces capture some clinical relationships well but do not yet provide balanced retrieval across heterogeneous evidence. 
We further provide MedEmb, an MLLM-based reference embedding model trained on the MedHEB ID split, as a reproducible baseline for unified 2D--3D--text medical retrieval. 
We release MedHEB with preprocessing scripts, evaluation code, access manifests, and MedEmb to support research on clinically meaningful medical embeddings.
\end{abstract}

\section{Introduction}\label{sec:intro}

\begin{figure}[h]
  \centering
  \includegraphics[width=\linewidth]{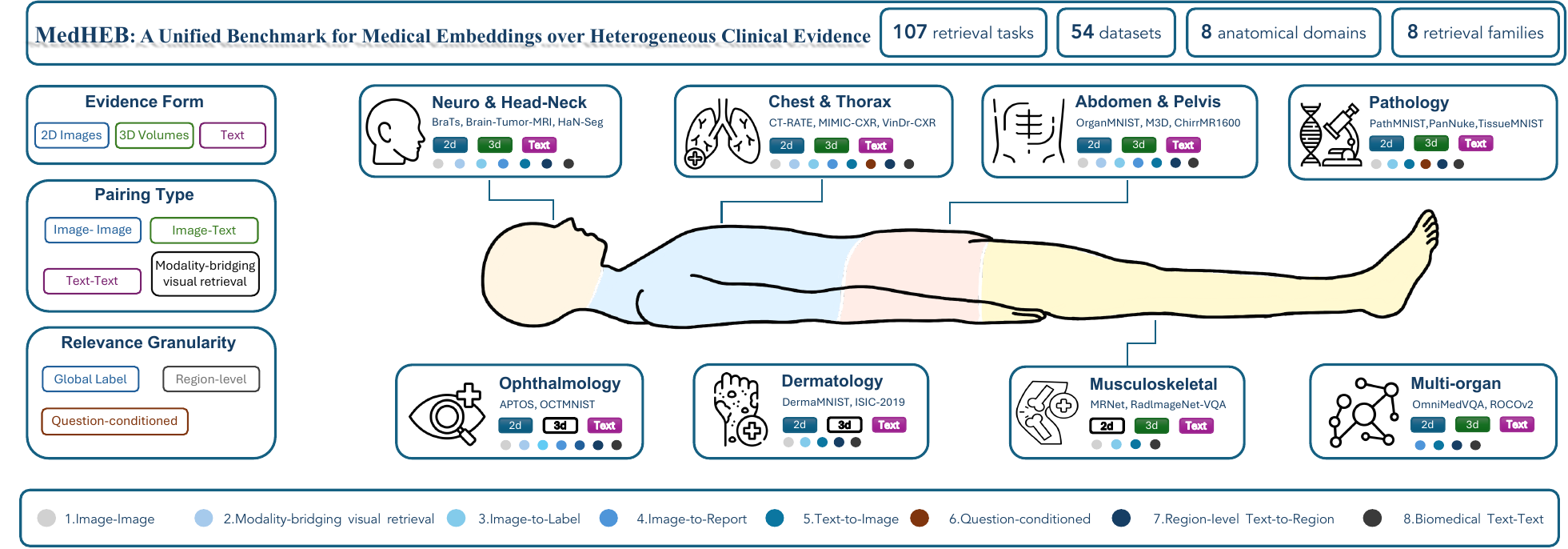}
  \caption{MedHEB design across clinical evidence axes and anatomical domains. The schematic organizes 54 datasets and 107 tasks by evidence form, pairing type, and relevance granularity.}
  \label{fig:medheb_overview}
  \vspace{-3mm}
\end{figure}

Radiologic interpretation is inherently comparative. In practice, radiologists interpret a current examination relative to prior studies, complementary modalities (e.g., CT or MRI), and morphologically similar cases, improving diagnostic confidence and accuracy~\cite{roelofs2007importance,hayward2016improving}. As foundation models enter radiology~\cite{zhang2024generalist,chen2024vision}, this process can be framed as retrieval over clinical evidence: the current study serves as the query, while prior exams, cross-modal data, reports, or localized findings form candidate evidence ranked by relevance. Shared embedding spaces make this formulation operational; however, “closeness” lacks a single clinical definition. Clinical similarity extends beyond visual resemblance: the same disease may manifest differently across modalities, and similar findings across patients may vary substantially in appearance~\cite{aaron2001review,muramatsu2018overview}. Relevant evidence also spans multiple forms, from cross-modal matches to report-level descriptions and region-level correspondences~\cite{sloan2024automated,wang2024survey,wang2025cross}. We therefore treat clinical similarity as task-conditioned, where relevance depends on the required relation and granularity. This reframes evaluation as whether a single embedding space preserves such similarity across changes in evidence form, pairing type, and relevance granularity.

Answering this question requires more than adding isolated retrieval tasks: it requires varying what ``relevance'' means in a controlled way. 
We organize clinical relatedness along three axes that a unified medical embedding space should support: \textbf{evidence form} (2D images, 3D volumes, and clinical text)~\cite{moor2023foundation, acosta2022multimodal}, \textbf{pairing type} (image--image, image--text, text--text, and modality-bridging visual matching)~\cite{choe2022content, chen2025survey}, and \textbf{relevance granularity} (global-label, question-conditioned, and localized-region)~\cite{fandeep}. 
Together, these axes specify what enters the embedding space, what relation defines a positive match, and at what scale relevance is judged. 
A benchmark confined to a single retrieval family can measure task-specific alignment, but cannot tell whether one embedding function forms a reusable clinical evidence space across changes in form, relation, and granularity.

Existing medical retrieval benchmarks standardize important but separate regimes: biomedical text retrieval in NFCorpus~\cite{boteva2016full}, BioASQ~\cite{krithara2023bioasq}, and R2MED~\cite{zhang2025r2med}; image--text and text--text retrieval in M3Retrieve~\cite{acharya2025m3retrieve}; and volumetric retrieval in 3D-MIR~\cite{abacha20233d} and BIMCV-R~\cite{chen2024bimcv}. 
However, they do not evaluate whether a single embedding function preserves clinically meaningful rankings as evidence form, pairing type, and relevance granularity change. 
This leaves clinically important relations underexplored, especially modality-bridging visual retrieval~\cite{liu2025m3ret} and region-level retrieval~\cite{lee2023region}.
To close this gap, we introduce \textbf{MedHEB}, a unified benchmark for evaluating medical embeddings over heterogeneous clinical evidence (Figure~\ref{fig:medheb_overview}). 
Rather than aggregating isolated retrieval tasks, MedHEB defines $107$ retrieval tasks from $54$ sources under a shared embedding-based ranking protocol. 
This design enables controlled evaluation across three axes (evidence form, pairing type, and relevance granularity), so models can be compared as clinically meaningful relations change within a single ranking protocol. 
MedHEB spans eight anatomical domains and integrates 2D images, 3D volumes, and clinical text, covering diverse retrieval families that instantiate these axes in concrete settings. 
By varying tasks while fixing the protocol, MedHEB directly tests whether a single embedding space preserves clinically meaningful rankings across heterogeneous evidence, rather than treating performance in isolated regimes such as image--text or text--text retrieval as a proxy for generalization. 
This formulation reveals structured differences across retrieval families: which relations are reliably captured, which remain weakly organized, and whether retrieval ability is balanced across families. 
We further provide \textbf{MedEmb}, a reference model for unified 2D--3D--text medical retrieval, as a baseline for the benchmark's target capability.

Our main contributions are as follows: \\
(1) We introduce \textbf{MedHEB}, a unified benchmark for medical embeddings over heterogeneous clinical evidence. MedHEB defines $107$ retrieval tasks from $54$ datasets across 2D images, 3D volumes, and clinical text, organized by evidence form, pairing type, and relevance granularity under a single embedding-based ranking protocol. Unlike prior benchmarks that evaluate isolated retrieval settings, MedHEB is designed as a controlled evaluation space over heterogeneous clinical relations.\\
(2) MedHEB extends medical embedding evaluation beyond conventional image--text and text--text alignment to underexplored relationships, notably \textbf{modality-bridging visual retrieval} and \textbf{region-level text-to-region retrieval}. Its analysis quantifies cross-family balance, showing which clinical relations current embeddings preserve reliably and which remain weakly captured. \\
(3) We release \textbf{MedHEB} with task indices, preprocessing scripts, evaluation code, and access manifests, and provide \textbf{MedEmb} as a reference MLLM-based baseline to anchor reproducible evaluation of unified 2D--3D--text medical retrieval.

\section{Related Work}

\textbf{Medical Retrieval Benchmarks.}~~
Existing medical retrieval benchmarks have standardized evaluation within important but separate regimes: biomedical text retrieval and question answering in NFCorpus~\cite{boteva2016full}, BioASQ~\cite{krithara2023bioasq}, and R2MED~\cite{zhang2025r2med}; image--text and text--text retrieval in M3Retrieve~\cite{acharya2025m3retrieve}; 3D image retrieval in 3D-MIR~\cite{abacha20233d}; and CT--report retrieval in BIMCV-R~\cite{chen2024bimcv}. 
What remains under-evaluated is whether a single embedding function preserves clinically meaningful rankings as evidence form, pairing type, and relevance granularity change under one protocol, especially for modality-bridging visual retrieval and region-level text-to-region retrieval. 
MedHEB addresses this gap by evaluating heterogeneous 2D image, 3D volume, and biomedical text under a common ranking protocol, with cross-regime placement quality and cross-family balance as primary evaluation targets.

\begin{table}[t]
  \caption{Coverage of representative medical retrieval benchmarks. I--T, T--T, I--I, and MB-VR denote image--text, text--text, image--image, and modality-bridging visual retrieval.}
  \label{tab:comparison-table}
  \centering
  \small
  \begin{tabular}{l c ccc ccccc}
    \toprule
    \rowcolor{TableHeader}
    \textbf{Benchmark} & \textbf{Year} & \multicolumn{3}{c}{\textbf{Evidence Form}} & \multicolumn{5}{c}{\textbf{Retrieval Setting}} \\
    \cmidrule(r){3-5} \cmidrule(r){6-10}
    \rowcolor{TableHeader}
    & & 2D & 3D & Text & I--T & T--T & I--I & MB-VR & Region \\
    \midrule
    BioASQ~\cite{krithara2023bioasq}        & 2015 &   &   & \checkmark &   & \checkmark &   &   &   \\
    NFCorpus~\cite{boteva2016full}          & 2016 &   &   & \checkmark &   & \checkmark &   &   &   \\
    3D-MIR~\cite{abacha20233d}              & 2023 &   & \checkmark &   &   &   & \checkmark &   &   \\
    BIMCV-R~\cite{chen2024bimcv}            & 2024 &   & \checkmark & \checkmark & \checkmark &   &   &   &   \\
    M3Retrieve~\cite{acharya2025m3retrieve} & 2025 & \checkmark &   & \checkmark & \checkmark & \checkmark &   &   &   \\
    R2MED~\cite{zhang2025r2med}             & 2025 &   &   & \checkmark &   & \checkmark &   &   &   \\
    \midrule
    \rowcolor{MedEmbRow}
    \textbf{MedHEB (ours)} & 2026 & \checkmark & \checkmark & \checkmark & \checkmark & \checkmark & \checkmark & \checkmark & \checkmark \\
    \bottomrule
  \end{tabular}
  \vspace{-4mm}
\end{table}

\textbf{Multimodal Embedding Models.}~~
Multimodal embedding models differ in how they compose heterogeneous evidence. 
Dual-tower models such as SigLIP~\cite{zhai2023sigmoid} and BLIP-2~\cite{li2023blip}, along with biomedical variants MedSigLIP~\cite{sellergren2025medgemma} and MedImageInsight~\cite{codella2024medimageinsight}, encode images and text separately, making them effective for global image--text matching but less direct for image--question prompts, report-conditioned retrieval, multi-slice volumes, or region-level grounding. 
MLLM-based embedding models, including VLM2Vec~\cite{meng2025vlm2vec}, RzenEmbed~\cite{jian2025rzenembed}, and Qwen3-VL-Embedding~\cite{li2026qwen3}, use a joint multimodal sequence encoder and can serialize mixed visual--textual evidence in one interface.

\begin{figure}[t]
  \centering
  \includegraphics[width=\linewidth]{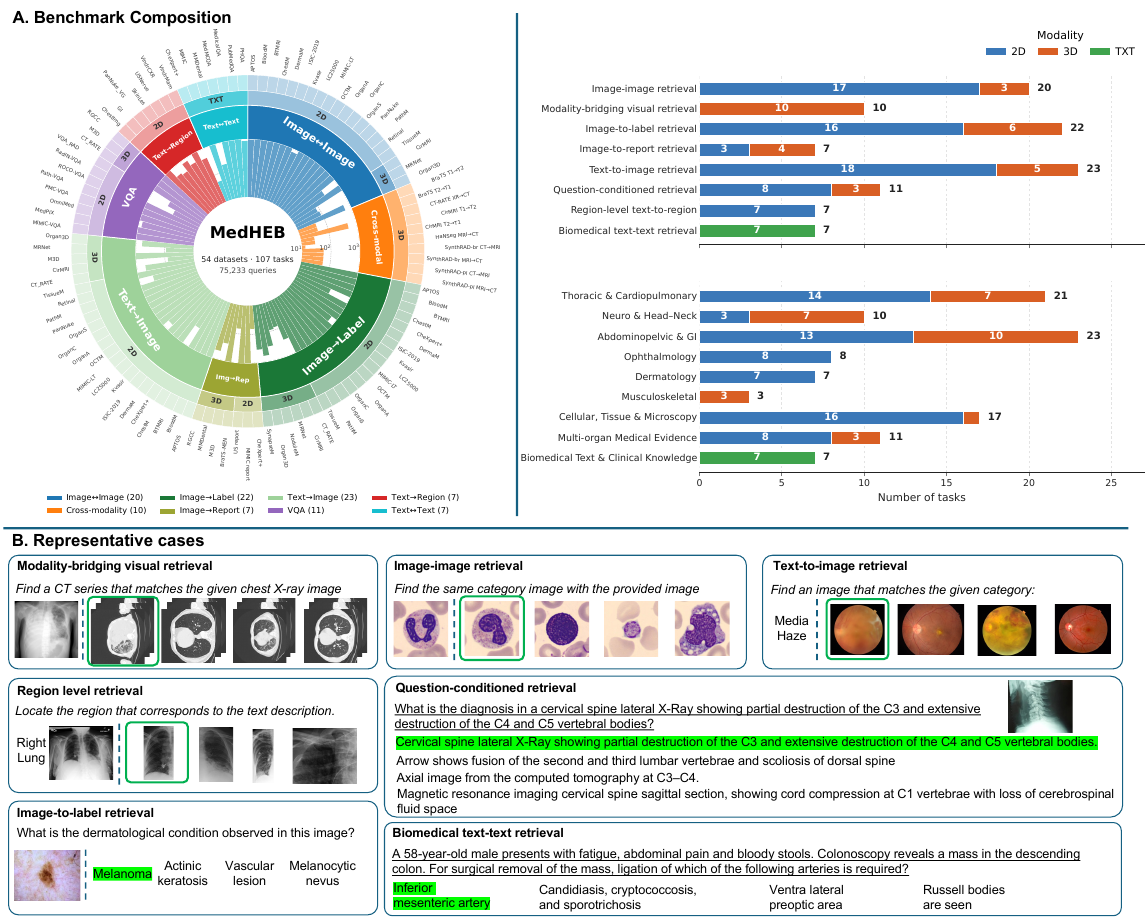}
  \caption{MedHEB task composition and representative retrieval examples. 
\textbf{A:} Hierarchical distribution of 107 tasks and 75{,}233 queries by evidence form, retrieval family, and source task, with side bars for family--modality and anatomy--modality counts. \textbf{B:} Query--candidate examples across modality-bridging, region-level, image--text, VQA, and biomedical text retrieval.}
  \label{fig:composition}
  \vspace{-4mm}
\end{figure}

\section{The MedHEB Benchmark}

MedHEB defines a controlled retrieval evaluation space over $54$ curated source datasets. Each of its $107$ tasks specifies a query, a candidate pool, and a clinically meaningful positive relation under one embedding-based ranking protocol: a model encodes both sides, ranks candidates by embedding similarity, and is evaluated by the placement of clinically relevant targets. The tasks span 2D images, 3D volumes, and biomedical text, organized into eight retrieval families that operationalize the three axes from Section~\ref{sec:intro}. MedHEB separates source familiarity from cross-dataset generalization through task-level ID/held-out-source splits. The benchmark contains $96$ in-distribution (ID) tasks and $11$ held-out-source tasks from six held-out datasets with non-overlapping sources. ID tasks follow standard train/test splits, while the held-out-source split introduces different anatomies, modalities, or reporting conventions, allowing assessment of transfer beyond the MedHEB training sources.


\begin{table}[h]
\vspace{-4pt}
\caption{Composition of MedHEB retrieval tasks by evidence form and split.}
\label{tab:medheb_composition}
\centering
\scriptsize
\setlength{\tabcolsep}{6pt}
\renewcommand{\arraystretch}{0.95}
\begin{tabular}{@{}lccc@{}}
\toprule
\textbf{Evidence Form} & \textbf{ID Tasks} & \textbf{Held-out-source Tasks} & \textbf{Total Tasks} \\
\midrule
2D images        & 60 & 9  & 69  \\
3D volumes       & 30 & 1  & 31  \\
Biomedical text  & 6  & 1  & 7   \\
\midrule
\textbf{Total}   & \textbf{96} & \textbf{11} & \textbf{107} \\
\bottomrule
\end{tabular}
\vspace{-6pt}
\end{table} 

\subsection{Constructing Heterogeneous Retrieval Tasks}\label{sec:construction}
MedHEB defines each task by a query form, candidate pool, and clinically meaningful positive relation under a shared ranking protocol, followed by common standardization, split construction, and quality control. The resulting $107$ tasks contain $75{,}233$ retrieval queries, whose distribution by evidence form, retrieval family, and anatomical coverage is summarized in Figure~\ref{fig:composition}.

\subsubsection{Unified Ranking Formulation}\label{sec:tasks}
All MedHEB tasks use the same retrieval protocol: given a query and a task-specific candidate pool, the evaluated model encodes both sides and ranks candidates by cosine similarity. 
Each query has one designated positive~\footnote{Multiple clinically relevant matches may exist, but most public datasets do not provide consistent multi-positive annotations. We therefore use a single-positive formulation for controlled comparison, and complement it with cross-source graded retrieval (Section~\ref{sec:graded-ood}) and clinician preference evaluation (Section~\ref{sec:human-eval}).}.
For label-retrieval tasks, we filter multi-label samples to avoid false negatives. \\
\textbf{Retrieval families.}~~
The families are defined by the clinical relation that makes a candidate relevant. The eight retrieval families instantiate complementary clinical relations, spanning visual similarity, cross-modal visual matching, image--text alignment, question-conditioned evidence, localized grounding, and biomedical semantics. Table~\ref{tab:retrieval_families} summarizes the query form, candidate pool, relevance definition, and representative sources for each family; per-task instruction templates and the full per-family dataset listings, and qualitative case studies are provided in Appendices~\ref{app:instructions}, \ref{app:family-details}, and~\ref{app:case-studies}.

\begin{table}[t]
\vspace{-4pt}
\caption{
MedHEB retrieval families. Each family follows the same ranking protocol but differs in query form, candidate pool. Full per-family dataset listings are provided in Appendix~\ref{app:family-details}.
}
\label{tab:retrieval_families}
\centering
\scriptsize
\setlength{\tabcolsep}{3.5pt}
\renewcommand{\arraystretch}{1.15}
\begin{tabularx}{\linewidth}{@{}p{0.18\linewidth}p{0.16\linewidth}X p{0.22\linewidth}@{}}
\toprule
\textbf{Family} & \textbf{Query} & \textbf{Candidates \& positive criterion} & \textbf{Representative sources} \\
\midrule

\textbf{Image--image} \newline (20 tasks)
& 2D image or 3D volume
& Same-source studies; positive if sharing the query's class, diagnosis, or anatomy label.
& e.g., Kvasir, LC25000, RFMiD~\cite{pogorelov2017kvasir,borkowski2019lung,pachade2021retinal} \\

\textbf{Modality-bridging visual} \newline (10 tasks)
& Image or volume from one modality
& Different-modality studies paired across MRI T1$\leftrightarrow$T2, CT$\leftrightarrow$MRI, or X-ray$\leftrightarrow$CT. Identifiers define positives only; not given to the model.
& e.g., BraTS, SynthRAD2023, HaN-Seg, CT-RATE~\cite{kazerooni2024brain,jha2025large,thummerer2023synthrad2023,podobnik2023han,hamamci2026generalist} \\

\textbf{Image-to-label} \newline (22 tasks)
& Image or volume
& Textual class labels; positive if matching the source label.
& e.g., CheXpert-Plus, MedMNIST~\cite{chambon2024chexpert,yang2023medmnist} \\

\textbf{Image-to-report} \newline (7 tasks)
& Image or volume
& Reports or captions; positive is the source-paired report.
& e.g., MIMIC-CXR, M3D, RadGen-CT~\cite{johnson2019mimic,chambon2024chexpert,li2024ultrasound,bai2024m3d,zhang2024radgenome,hamamci2026generalist,labella2023asnr,lei2024autorg,wang2025mmdental} \\

\textbf{Text-to-image} \newline (23 tasks)
& Class label or short clinical description
& Same-source images matching the queried label or description.
& Reuses image-to-label and image-to-report image sources. \\

\textbf{Question-conditioned} \newline (11 tasks)
& Image--question pair
& Dataset-specific answer set; positive if matching the ground-truth answer.
& e.g., MIMIC-CXR-VQA, OmniMedVQA, Path-VQA~\cite{PhysioNet-medical-cxr-vqa-dataset-1.0.0,siragusa2025medpix,hu2024omnimedvqa,zhang2023pmc,he2020pathvqa,pelka2018roco,butsanets2025radimagenet,lau2018dataset} \\

\textbf{Region-level retrieval} \newline (7 tasks)
& Full image with a region-level text description
& Source-provided region crops matching the text description.
& e.g., Chest-ImaGenome, PanNuke, VinDr-CXR~\cite{PhysioNet-chest-imagenome-1.0.0,jha2019kvasir,gamper2019pannuke,kucs2024medsegbench,codella2019skin,tschandl2018ham10000,ultrasound-nerve-segmentation,PhysioNet-vindr-cxr-1.0.0,PhysioNet-vindr-mammo-1.0.0} \\

\textbf{Biomedical text--text} \newline (7 tasks)
& Biomedical text passage
& Same-source passages: QA, findings$\to$impression, or diagnosis$\to$treatment.
& e.g., MedMCQA, PubMedQA, MIMIC-CXR~\cite{pal2022medmcqa,jin2019pubmedqa,ben2019question,johnson2019mimic,chambon2024chexpert} \\

\bottomrule
\end{tabularx}
\vspace{-6pt}
\end{table}

\subsubsection{Dataset Selection and Clinical Coverage}\label{sec:datasets}

We select source datasets to support controlled variation along the MedHEB axes. Included sources must contribute at least one benchmark role (evidence-form diversity, clinically meaningful pairing relations, target-granularity variation, anatomical breadth, or dataset-level held-out-source evaluation) and be reproducible through direct redistribution or an official access route.
For restricted datasets, we release manifests and preprocessing scripts so authorized users can reconstruct the benchmark under their own credentials.
MedHEB covers more than 25 anatomical sites across eight clinical groups: neuro/head-neck (e.g., BraTS, HaN-Seg)~\cite{kazerooni2024brain,podobnik2023han,wang2025mmdental}, chest/thorax (e.g., CT-RATE, MIMIC-CXR)~\cite{hamamci2026generalist,johnson2019mimic,holste2022long,PhysioNet-vindr-cxr-1.0.0}, abdomen/pelvis (e.g., OrganMNIST, M3D)~\cite{yang2023medmnist,bai2024m3d,jha2025large}, musculoskeletal~\cite{bien2018deep}, dermatology~\cite{hernandez2024bcn20000}, ophthalmology~\cite{aptos2019-blindness-detection,yang2023medmnist}, pathology~\cite{yang2023medmnist,gamper2019pannuke}, and multi-region resources~\cite{hu2024omnimedvqa,pelka2018roco}.
This breadth reduces single-domain bias and supports evaluation across heterogeneous clinical evidence. Full source, access, and dataset-to-task mappings are provided in the Appendix~\ref{app:family-details}.

\subsubsection{Standardization, Splits, and Quality Control}\label{sec:preprocessing}

\textbf{Input standardization.}~~
We standardize inputs only where needed for comparable retrieval while preserving source evidence. 3D CT/MRI studies are represented by 64 uniformly sampled frames; CT$\leftrightarrow$X-ray tasks in CT-RATE use DeepDRR-synthesized radiographs from paired CT volumes~\cite{unberath2018deepdrr}; and region candidates are cropped from source-provided bounding boxes. \\
\textbf{Pairing and false-negative control.}~~
MB-VR positives are constructed only from within-dataset patient or study identifiers, with no cross-dataset patient linking. Identifiers define high-confidence evaluation pairs and are never provided to models, so MB-VR evaluates image-based cross-modal visual matching rather than identifier lookup. For image-to-label retrieval, we retain only single-positive samples to ensure non-target labels are valid negatives. These rules reduce false-negative bias while keeping tasks reconstructable from public annotations. \\
\textbf{Splits and held-out-source evaluation.}~~
We follow official train/validation/test partitions when available and otherwise use a 7:3 train--test split. Held-out-source tasks are drawn from held-out sources not used in ID construction, including CheXpert-Plus, RFMiD, LC25000, OmniMedVQA, and BraTS-MEN. Held-out sources are excluded from MedHEB training and may differ from ID sources in image characteristics, label distributions, or reporting conventions. Results on these sources therefore measure cross-dataset transfer.

\section{MedEmb Reference Baseline}\label{sec:baseline}

\begin{figure}[t]
  \centering
  \includegraphics[width=0.8\linewidth]{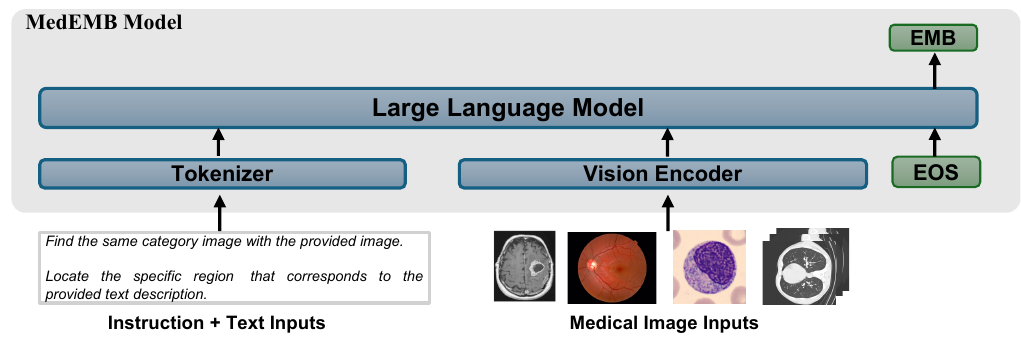}
  \caption{MedEmb reference model. Text, 2D images, 3D volumes, and mixed image--text inputs are serialized into one MLLM, and the final-token hidden state is used as the retrieval embedding.}
  \label{fig:medemb_arch}
  \vspace{-4mm}
\end{figure}

MedEmb is designed as a reproducible reference baseline for MedHEB (Figure~\ref{fig:medemb_arch}), illustrating one implementation of unified 2D--3D--text embedding under the benchmark's retrieval formulation. 
It adapts Qwen3-VL-Embedding~\cite{li2026qwen3} with a single sequence interface: inputs are formatted by the model-specific chat template into a tokenized sequence $\tilde{x}$ containing text prompts, role markers, image placeholders, and special tokens. 
The MLLM $f_\theta$ encodes $\tilde{x}$, and the retrieval embedding is read from the last-layer hidden state at the final \texttt{<|endoftext|>} token, $\mathbf{h}_x=f_\theta(\tilde{x})\in\mathbb{R}^d$~\cite{jiangvlm2vec,li2026qwen3}. 
This interface supports text, 2D images, sampled 3D volumes, and interleaved image--text inputs without task-specific fusion modules. 
MedEmb-2B/8B are initialized from Qwen3-VL-Embed-2B/8B and fine-tuned on the MedHEB ID training split for one epoch using LoRA adapters ($r=8$, $\alpha=32$), AdamW with cosine decay and warmup, peak learning rate $1\mathrm{e}{-4}$, contrastive temperature $\tau=0.01$, false-negative margin $\delta=0.1$, token-budgeted dynamic batching, and $8$ NVIDIA H100 GPUs.\\
\textbf{Instruction-conditioned contrastive objective.}~~
Each training instance is an instruction-prefixed query--target pair $(\mathbf{q}_i,\mathbf{t}_i^+)$. 
Both sides are encoded by the same backbone, $\mathbf{h}_{q_i}=f_\theta(\mathbf{q}_i)$ and $\mathbf{h}_{t_j^+}=f_\theta(\mathbf{t}_j^+)$, and matched by temperature-scaled cosine similarity,
$\phi(\mathbf{h}_q,\mathbf{h}_t)=\exp\!\left(\tfrac{1}{\tau}\cos(\mathbf{h}_q,\mathbf{h}_t)\right)$. 
For a batch of size $B$, other targets in the batch serve as negatives for $\mathbf{q}_i$ under the standard InfoNCE loss
\begin{equation}
\mathcal{L}_{\text{InfoNCE}} \;=\; -\frac{1}{B}\sum_{i=1}^{B} \log
    \frac{\phi(\mathbf{h}_{q_i}, \mathbf{h}_{t_i^+})}
         {\sum_{j=1}^{B} \phi(\mathbf{h}_{q_i}, \mathbf{h}_{t_j^+})}.
\end{equation} 
\textbf{False-negative masking.}~~
Medical retrieval batches may contain candidates clinically equivalent to the anchor, such as another normal chest X-ray. 
Pushing such samples away would penalize meaningful similarity, especially when they score above the annotated positive. 
We therefore mask in-batch candidates whose similarity exceeds the ground-truth positive by margin $\delta$ before computing InfoNCE, yielding $\mathcal{L}_{\text{MedEmb}}$; the full objective is in Appendix~\ref{app:medemb-details}. \\
\textbf{Token-budgeted dynamic batching.}~~
MedHEB inputs vary from short labels to long reports, volumetric studies, and multi-image sequences, making fixed-size batching padding-inefficient. 
We use token-budgeted dynamic batching, where batch size varies while padded token cost is bounded by $T=2B_{\text{nom}}L_{\max}$, matching the static-batching budget at nominal batch size $B_{\text{nom}}$ and maximum length $L_{\max}$. 
The formal batch constraint and online best-fit packing are in Appendix~\ref{app:medemb-details}.

\section{Benchmark Experiments}\label{sec:experiments}

\subsection{Evaluation Protocol, Metrics, and Models}\label{sec:metrics}

All models are evaluated under the same MedHEB ranking protocol: for each task, a model embeds the query and task-specific candidate pool, ranks candidates by cosine similarity, and is scored by the placement of clinically relevant targets. 
We report Hit@1 and MRR@5 as primary metrics, with Hit@5 in Appendix~\ref{app:hit5}. 
Hit@1 measures whether the top-ranked candidate is clinically relevant, while MRR@5 averages the reciprocal rank of the first relevant candidate within the top five, crediting near-misses that Hit@1 ignores. 
To assess whether retrieval ability is balanced across clinical relations, we also report cross-family consistency using the coefficient of variation (CV) of family-level Hit@1, where lower CV indicates more even performance across retrieval families.

We evaluate eight publicly available embedding models from two architectural families. 
\textbf{Dual-tower encoders} include \textbf{SigLIP}~\cite{zhai2023sigmoid}, \textbf{BLIP-2}~\cite{li2023blip}, \textbf{MedSigLIP}~\cite{sellergren2025medgemma}, and \textbf{MedImageInsight}~\cite{codella2024medimageinsight}. 
Because they encode images and text separately, mixed-evidence inputs are embedded field-wise, mean-pooled, and $\ell_2$-normalized; volumetric or multi-image inputs use at most $K{=}64$ uniformly sampled slices or frames before pooling. 
\textbf{MLLM-based models} include \textbf{VLM2Vec-V2.0}~\cite{meng2025vlm2vec}, \textbf{RzenEmbed}~\cite{jian2025rzenembed}, and \textbf{Qwen3-VL-Embed-2B/8B}~\cite{li2026qwen3}. 
These models serialize text, role markers, image placeholders, and sampled volumetric slices through model-specific chat templates into one multimodal token stream, using the final-token hidden state as the embedding.



\subsection{Main Results}\label{sec:main-results}
Table~\ref{tab:main_results_medical} reveals substantial cross-family imbalance in current medical embedding spaces. 
Across public baselines, strong performance in one retrieval family does not imply uniformly strong performance across clinical relations. 
For example, Qwen3-VL-Embed-8B achieves high biomedical text--text retrieval performance, but scores much lower on modality-bridging visual retrieval and region-level text-to-region retrieval. The same pattern appears under both Hit@1 and MRR@5.

The family-wise and ID/held-out-source breakdowns further show why aggregate averages are insufficient for evaluating medical embeddings.
Established families such as image--image, image-to-label, text-to-image, and text--text retrieval are comparatively tractable, whereas modality-bridging visual retrieval, and region-level localization remain less consistently captured. 
The ID/held-out-source split adds an orthogonal dimension by separating retrieval within source distributions from transfer to held-out sources with different images, labels, anatomies, or reporting conventions.
Together, these views show that current embedding spaces are uneven across relation type, target granularity, and distribution shift, rather than uniformly capable under a single aggregate score.
MedEmb also leads an instruction-free comparison on $49$ tasks (Appendix~\ref{app:instruction-free}).

\begin{table}[t]
\centering
\caption{
Main MedHEB retrieval results. Entries are Hit@1 with MRR@5 in parentheses (\%). Family columns average retrieval families; summary columns average ID, held-out-source, and all tasks. Abbreviations: I = image, T = text, MB = modality-bridging, VR = visual retrieval, Lab = label, Rep = report, VQA = question-conditioned retrieval, VG = region-level text-to-region.
}
\label{tab:main_results_medical}
\footnotesize
\setlength{\tabcolsep}{2pt}
\renewcommand{\arraystretch}{1.0}
\newcommand{\scorecell}[2]{\makecell{#1\\[-3pt]\scriptsize \textcolor{MetricGray}{(#2)}}}
\newcommand{\bestscorecell}[2]{\cellcolor{BestCell}\makecell{\textbf{#1}\\[-3pt]\scriptsize \textbf{\textcolor{MetricGray}{(#2)}}}}
\begin{tabular}{lcccccccccccc}
\toprule
\rowcolor{TableHeader}
\textbf{Model}
& \multicolumn{8}{c}{\textbf{Retrieval Families}}
& \multicolumn{3}{c}{\textbf{Summary}} \\
\cmidrule(r){2-9} \cmidrule(r){10-12}
\rowcolor{TableHeader}
& I-I & MB-VR & I-Lab & I-Rep & T-I & VQA & VG & T-T
& ID & \makecell{Held-out-\\source} & All \\
\midrule
\rowcolor{TableSubHeader}
\textbf{\# of Tasks $\rightarrow$}
& 20 & 10 & 22 & 7 & 23 & 11 & 7 & 7
& 96 & 11 & 107 \\
\midrule
\rowcolor{TableGroup}
SigLIP~\cite{zhai2023sigmoid}
& \scorecell{31.39}{48.13} & \scorecell{6.78}{11.19} & \scorecell{12.19}{22.66} & \scorecell{5.00}{8.05} & \scorecell{20.91}{36.13} & \scorecell{2.64}{4.86} & \scorecell{2.65}{4.15} & \scorecell{9.45}{13.88}
& \scorecell{14.53}{24.09} & \scorecell{18.01}{29.78} & \scorecell{14.89}{24.67} \\

\rowcolor{TableStripe}
BLIP-2~\cite{li2023blip}
& \scorecell{29.68}{46.88} & \scorecell{7.23}{12.02} & \scorecell{7.05}{16.51} & \scorecell{5.65}{11.41} & \scorecell{12.85}{27.70} & \scorecell{3.38}{5.46} & \scorecell{2.70}{4.16} & \scorecell{7.91}{11.75}
& \scorecell{11.27}{20.73} & \scorecell{16.91}{29.04} & \scorecell{11.85}{21.58} \\

MedImageInsight~\cite{codella2024medimageinsight}
& \scorecell{44.74}{59.48} & \scorecell{10.71}{18.51} & \scorecell{14.25}{25.55} & \scorecell{9.90}{16.71} & \scorecell{26.13}{41.11} & \scorecell{10.22}{13.88} & \scorecell{7.05}{9.14} & \scorecell{26.84}{34.57}
& \scorecell{21.28}{31.47} & \scorecell{26.57}{39.77} & \scorecell{21.83}{32.32} \\

\rowcolor{TableStripe}
MedSigLIP~\cite{sellergren2025medgemma}
& \scorecell{19.73}{37.22} & \scorecell{3.71}{7.34} & \scorecell{13.50}{23.35} & \scorecell{4.86}{9.63} & \scorecell{24.18}{39.72} & \scorecell{6.58}{9.62} & \scorecell{0.46}{1.05} & \scorecell{11.42}{17.24}
& \scorecell{13.47}{23.04} & \scorecell{16.53}{30.36} & \scorecell{13.78}{23.80} \\

RzenEmbed~\cite{jian2025rzenembed}
& \scorecell{31.04}{47.88} & \scorecell{8.62}{14.77} & \scorecell{13.86}{24.73} & \scorecell{7.99}{15.68} & \scorecell{17.25}{31.95} & \scorecell{4.76}{7.85} & \scorecell{7.96}{10.12} & \scorecell{58.47}{64.01}
& \scorecell{18.21}{28.46} & \scorecell{21.30}{33.37} & \scorecell{18.52}{28.97} \\

\rowcolor{TableStripe}
VLM2Vec-V2.0~\cite{meng2025vlm2vec}
& \scorecell{26.27}{43.17} & \scorecell{6.18}{12.27} & \scorecell{14.33}{27.25} & \scorecell{7.22}{14.40} & \scorecell{15.60}{30.50} & \scorecell{17.57}{22.84} & \scorecell{3.84}{5.45} & \scorecell{39.71}{46.73}
& \scorecell{16.67}{27.69} & \scorecell{19.08}{31.50} & \scorecell{16.91}{28.08} \\

Qwen3-VL-Embed-2B~\cite{li2026qwen3}
& \scorecell{33.04}{49.73} & \scorecell{10.60}{17.77} & \scorecell{18.90}{32.65} & \scorecell{9.83}{17.70} & \scorecell{21.46}{36.41} & \scorecell{21.04}{28.16} & \scorecell{4.10}{5.47} & \scorecell{57.32}{62.24}
& \scorecell{22.16}{33.55} & \scorecell{25.33}{37.75} & \scorecell{22.49}{33.98} \\

\rowcolor{TableStripe}
Qwen3-VL-Embed-8B~\cite{li2026qwen3}
& \scorecell{34.69}{51.31} & \scorecell{8.99}{15.67} & \scorecell{20.70}{34.35} & \scorecell{11.52}{18.83} & \scorecell{20.08}{35.22} & \scorecell{24.58}{32.76} & \scorecell{18.71}{22.05} & \scorecell{64.72}{70.01}
& \scorecell{24.32}{35.84} & \scorecell{27.41}{40.41} & \scorecell{24.64}{36.31} \\

\rowcolor{TableGroup}
\rowcolor{MedEmbRow}
\textbf{MedEmb-2B}
& \scorecell{52.03}{65.41} & \scorecell{14.49}{25.24} & \scorecell{56.43}{71.03} & \scorecell{25.76}{35.46} & \scorecell{52.75}{63.74} & \scorecell{33.47}{42.97} & \scorecell{33.47}{39.49} & \scorecell{65.81}{72.45}
& \scorecell{45.63}{57.16} & \bestscorecell{45.77}{55.17} & \scorecell{45.64}{56.95} \\

\rowcolor{MedEmbRow}
\textbf{MedEmb-8B}
& \bestscorecell{54.35}{67.47} & \bestscorecell{20.30}{30.70} & \bestscorecell{58.83}{72.51} & \bestscorecell{28.97}{41.12} & \bestscorecell{55.70}{66.53} & \bestscorecell{36.88}{46.97} & \bestscorecell{44.17}{50.31} & \bestscorecell{70.64}{76.96}
& \bestscorecell{49.91}{61.22} & \scorecell{44.22}{54.59} & \bestscorecell{49.32}{60.53} \\
\bottomrule
\end{tabular}
\vspace{-4mm}
\end{table}

\subsection{Analysis}\label{sec:analysis}

Since all MedHEB tasks are evaluated in a shared embedding space, we interpret the following analyses as measurements of how each model organizes clinically relevant evidence around the query.

\begin{figure}[h]
  \centering
  \includegraphics[width=\linewidth]{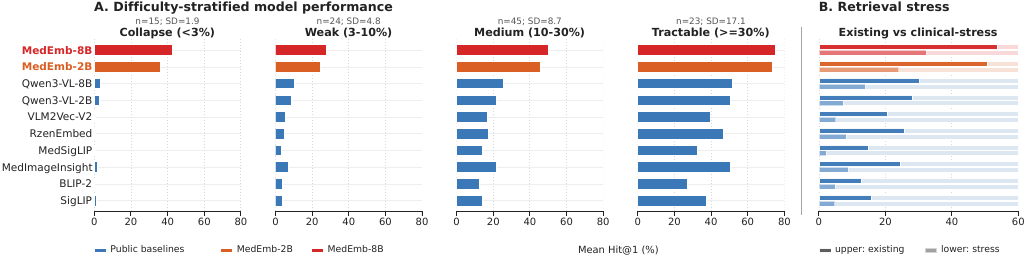}
  \caption{Benchmark difficulty and cross-relation evaluation. 
\textbf{A:} Tasks are grouped by public-baseline mean Hit@1 into Collapse, Weak, Medium, and Tractable regimes. 
\textbf{B:} Paired model scores compare established families (I-I, I-Lab, I-Rep, T-I, T-T) with underexplored relation families (MB-VR, VG).}
  \vspace{-3mm}
  \label{fig:difficulty_profile}
\end{figure}

\textbf{Task separability and benchmark discriminativeness.}~~
Figure~\ref{fig:difficulty_profile}A stratifies the $107$ MedHEB tasks by mean public-baseline Hit@1 into four buckets: Collapse ($<3\%$, $n=15$), Weak ($3$--$10\%$, $n=24$), Medium ($10$--$30\%$, $n=45$), and Tractable ($\geq30\%$, $n=23$). 
Because this stratification uses public baselines, it reflects the benchmark's difficulty structure rather than any proposed model. 
MedHEB is neither saturated nor uniformly unsolved: Collapse and Weak tasks mark low-signal settings where current embeddings rarely place the correct target near the query, while Medium and Tractable tasks provide increasing model separation. 
The between-baseline standard deviation rises from $1.9$ to $4.8$, $8.6$, and $17.1$ across buckets, showing that MedHEB contains both underdeveloped clinical retrieval settings and higher-signal tasks that meaningfully distinguish embedding models.

\textbf{Transfer across retrieval relations.}~~
To quantify cross-relation transfer, we compare established families covered by prior benchmarks (image--image, image-to-label, image-to-report, text-to-image, biomedical text--text) with underexplored clinical-relation families (modality-bridging visual retrieval and region-level text-to-region retrieval). Figure~\ref{fig:difficulty_profile}B shows a consistent $7.7$--$26.6$ Hit@1 drop from established to underexplored relations across all models. Thus, conventional image--text, text--text, or image--image neighborhoods do not necessarily preserve cross-modal visual matches or localized evidence, and MedHEB measures this cross-relation gap beyond aggregate retrieval accuracy.

\begin{figure}[h]
  \centering
  \includegraphics[width=\linewidth]{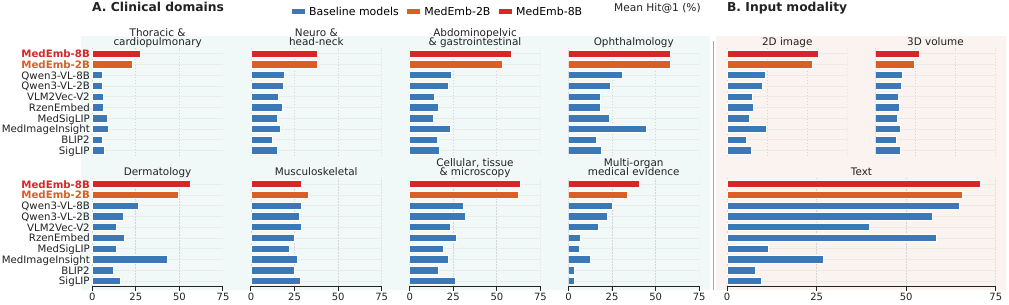}
  \caption{Retrieval difficulty across anatomical domains and evidence forms.}
  \label{fig:domain_breakdown}
  \vspace{-3mm}
\end{figure}

\textbf{Domain- and modality-dependent embedding behavior.}~~
Figure~\ref{fig:domain_breakdown} shows that retrieval difficulty tracks domain-specific invariances. 
Standardized or morphology-driven settings (e.g., ophthalmology, dermatology, abdominal imaging, and tissue microscopy) are more tractable because relevant candidates cluster around repeatable visual patterns, anatomical priors, or localized texture~\cite{sahlsten2019deep,reddy2024artificial,dougan2024histopathology}. 
By contrast, thoracic/cardio, musculoskeletal, neuro/head-neck, and multi-organ tasks often require embeddings to capture subtle findings, sparse abnormalities, cross-sectional anatomy, cross-modal shifts, or report-level semantics~\cite{diagnostics13040743}. 
The modality breakdown highlights 3D retrieval as a bottleneck: unlike many 2D tasks with standardized global views, volumes require spatial continuity, small cross-slice findings, acquisition-specific contrast, and cross-modal anatomical structure~\cite{zhang2024single}. 
The 2D--3D gap therefore reflects weaker organization of volumetric evidence, not merely longer inputs.

\begin{figure}[h]
  \centering
  \includegraphics[width=\linewidth]{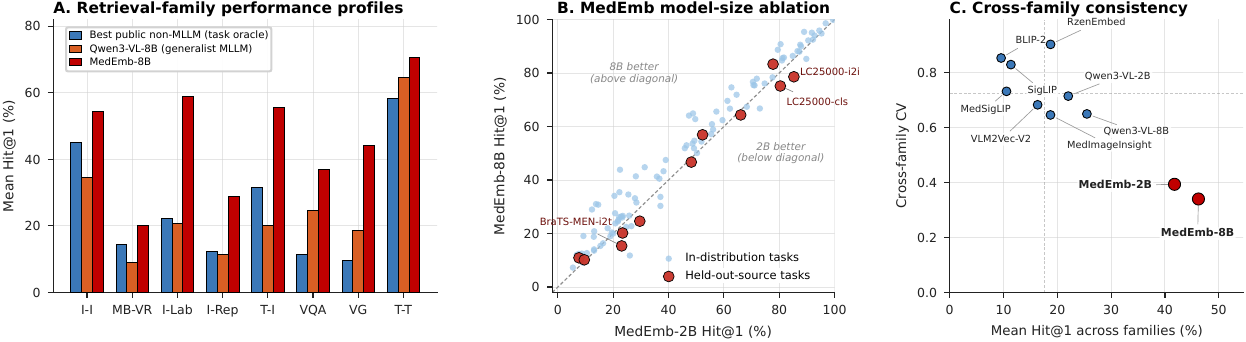}
  \caption{Capability, model-size ablation, and consistency profiles. \textbf{A:} Family-level Hit@1 for a strong non-MLLM baseline, Qwen3-VL-8B, and MedEmb. \textbf{B:} Per-task MedEmb-2B vs MedEmb-8B scores for ID and held-out-source tasks. \textbf{C:} Mean family Hit@1 vs cross-family CV for accuracy–balance trade-off.}
  \label{fig:architecture_scaling}
  \vspace{-4mm}
\end{figure}

\textbf{Embedding capability and model-size ablation.}~~
Figure~\ref{fig:architecture_scaling}A separates relations that current embeddings organize well from those that remain weakly represented. 
Image--image and biomedical text--text retrieval are comparatively strong, while larger gaps appear when embeddings must connect visual evidence to clinical terminology, report phrasing, disease hierarchy, or localized findings, as in image-to-label, image-to-report, text-to-image, question-conditioned, and region-level retrieval. 
The MedEmb-2B/8B comparison provides a model-size ablation: the 8B model improves many ID tasks but less consistently improves held-out-source tasks (Figure~\ref{fig:architecture_scaling}B), suggesting that larger model size helps alignment within familiar sources but does not guarantee robust retrieval under shifts in label ontology, anatomy, staining style, or reporting convention.
These results motivate training signals that bind clinical relations across datasets, modalities, and granularities.
On a fixed training subset, mixed-task training achieves higher overall Hit@1 than training on individual task groups (Appendix~\ref{app:task-group-ablation}).

\textbf{Cross-family consistency of embedding geometry.}~~
A unified clinical embedding should not concentrate performance in a few retrieval families. 
We measure this balance using the coefficient of variation of family-level Hit@1, $\mathrm{CV}(m)=\frac{\mathrm{std}_{k}(h_{m,k})}{\mathrm{mean}_{k}(h_{m,k})+\epsilon}$,
where $h_{m,k}$ is model $m$'s task-macro Hit@1 on family $k$, lower CV indicates more even performance. 
Since CV measures dispersion rather than accuracy, Figure~\ref{fig:architecture_scaling}C plots it against mean family Hit@1. 
Public embeddings show high variability (CV $0.65$--$0.90$), meaning aggregate scores are concentrated in a subset of families. 
MedEmb serves as a reference point with lower CV ($0.34$--$0.40$) and higher mean Hit@1, indicating more balanced family-level retrieval; per-model statistics are in Appendix~\ref{app:consistency}.

\subsection{Graded-Relevance Evaluation}\label{sec:graded-relevance}
To evaluate retrieval beyond a single annotated positive, we extend 12 MedHEB datasets with graded relevance covering severity, laterality, diagnosis, and cell/tissue hierarchy.
Relevance is derived from source annotations and published clinical hierarchies, with partial credit for clinically related candidates.
Detailed grading rules and weights are provided in Appendix~\ref{app:graded-protocol}.
Both MedEmb variants outperform the baselines across all four relation types (Table~\ref{tab:graded_relevance}), supporting the benchmark conclusions beyond single-positive matching.

\begin{table}[!htbp]
\centering
\caption{Graded-relevance retrieval performance (NDCG@5, \%).}
\label{tab:graded_relevance}
\footnotesize
\setlength{\tabcolsep}{5pt}
\renewcommand{\arraystretch}{0.95}
\begin{tabular}{@{}lrrrrr@{}}
\toprule
Model & Severity & Laterality & Diagnosis & Cell/tissue & All \\
\midrule
BLIP-2 & 67.31 & 32.02 & 48.85 & 32.61 & 47.47 \\
MedSigLIP & 70.06 & 29.78 & 51.40 & 40.49 & 50.72 \\
SigLIP & 69.63 & 34.82 & 52.37 & 40.21 & 51.55 \\
RzenEmbed & 67.62 & 31.66 & 51.97 & 45.92 & 51.85 \\
VLM2Vec-V2.0 & 71.29 & 30.40 & 52.04 & 44.45 & 52.23 \\
MedImageInsight & 70.61 & 36.56 & 61.76 & 43.64 & 56.90 \\
Qwen3-VL-Embed-2B & 79.75 & 41.41 & 59.53 & 46.64 & 59.14 \\
Qwen3-VL-Embed-8B & 82.44 & 41.46 & 60.56 & 43.45 & 59.50 \\
\midrule
\rowcolor{MedEmbRow}
MedEmb-2B & 86.74 & 89.07 & \textbf{77.57} & 85.39 & 82.62 \\
\rowcolor{MedEmbRow}
MedEmb-8B & \textbf{86.96} & \textbf{92.03} & 77.56 & \textbf{86.95} & \textbf{83.34} \\
\bottomrule
\end{tabular}
\end{table}

\subsection{Cross-Source Graded Retrieval}\label{sec:graded-ood}
\begin{figure}[h]
\centering
\begin{minipage}[c]{0.39\linewidth}
  \centering
  \scriptsize
  \setlength{\tabcolsep}{2pt}
  \renewcommand{\arraystretch}{0.95}
  \captionof{table}{Results of cross-source graded retrieval.}
  \resizebox{\linewidth}{!}{%
  \begin{tabular}{@{}lcccc@{}}
    \toprule
    Model & NDCG@5 & NDCG@10 & Hit@1 & Hit@5 \\
    \midrule
    Qwen3-VL-Embed-2B & 0.484 & 0.652 & 0.547 & 0.897 \\
    Qwen3-VL-Embed-8B & 0.460 & 0.633 & 0.467 & 0.897 \\
    \midrule
    MedEmb-2B & 0.543 & 0.689 & 0.627 & 0.937 \\
    MedEmb-8B & \textbf{0.566} & \textbf{0.694} & \textbf{0.637} & \textbf{0.940} \\
    \bottomrule
  \end{tabular}
  }
  \label{tab:graded_ood}
\end{minipage}\hfill
\begin{minipage}[c]{0.58\linewidth}
  \centering
  \includegraphics[width=\linewidth]{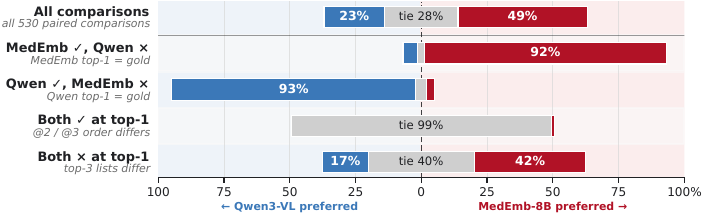}
  \captionof{figure}{Human evaluation of MedEmb versus Qwen3-VL-Embed top-3 lists, stratified by top-1 correctness.}
  \label{fig:human_eval}
  \vspace{-4mm}
\end{minipage}
\end{figure}

To complement MedHEB's single-positive protocol, we introduce a graded held-out-source retrieval setting where relevance is defined softly by shared clinical attributes, with queries and candidate pools drawn from different datasets. Specifically, we construct a graded image-to-image retrieval task with queries from MIMIC-CXR and candidates from CheXpert-Plus, two chest X-ray datasets differing in source, acquisition protocol, and post-processing; MedEmb does not see CheXpert-Plus during MedHEB training. Both queries and candidates are mapped to the shared 14-label CheXpert vocabulary, and relevance is defined by the \emph{Jaccard overlap} of their multi-label annotations, yielding scores in $[0,1]$. We report NDCG@$K$ and Hit@$K$. As shown in Table~\ref{tab:graded_ood}, MedEmb consistently outperforms the strongest Qwen3-VL-Embed baselines across all metrics without further training, indicating that the learned embedding space transfers across data sources under distribution shift.

\subsection{Human Evaluation}\label{sec:human-eval}
Automatic metrics based on exact matches do not fully capture clinical retrieval utility: clinicians inspect top-ranked neighbors for differential diagnosis and prior-case comparison, where plausible alternatives matter even when the annotated gold is not retrieved. 
We therefore run a single-blind head-to-head study in which a board-certified clinician compares the top-3 lists from MedEmb-8B and Qwen3-VL-Embed-8B on $530$ queries stratified across all $107$ MedHEB tasks, with model identity, gold answer, and stratum hidden (Appendix~\ref{app:human-protocol}).
MedEmb-8B is preferred in $49.1\%$ of comparisons versus $23.0\%$ for Qwen3-VL-Embed-8B, with $27.9\%$ ties (Figure~\ref{fig:human_eval}). When both models miss the gold at top-1, MedEmb-8B is still preferred in $42.3\%$ of cases versus $17.3\%$ for Qwen, suggesting that ranked neighborhoods contain clinically meaningful signal beyond exact top-1 accuracy. 
This supports MedHEB's use of ranked retrieval as an evaluation signal for clinical evidence retrieval.

\section{Conclusion}\label{sec:conclusion}
We introduced MedHEB, a unified benchmark that shifts medical embedding evaluation from isolated retrieval scores to the organization of clinical relatedness in a shared space. 
By fixing the ranking protocol while varying evidence form, pairing type, and relevance granularity, MedHEB tests whether one embedding function preserves clinically meaningful neighborhoods across heterogeneous evidence, including modality-bridging visual and region-level text-to-region retrieval. 
Experiments show that current embeddings remain uneven across retrieval families, motivating evaluation of family-level placement quality and cross-family balance rather than a single aggregate score. 
We release MedHEB with task indices, preprocessing and evaluation code, and access manifests, together with MedEmb as a reference baseline, to support reproducible work on unified clinical embeddings.

\medskip

\newpage
{\small
\bibliographystyle{IEEEtran}
\bibliography{main}
}

\appendix
\section{Supplementary Material}

\FloatBarrier
\subsection{MedEmb Training Details}\label{app:medemb-details}

\paragraph{False-negative mask.}
We define the mask factor $m_{ij}$ with margin $\delta$ as
\begin{equation}
m_{ij} \;=\;
\begin{cases}
0, & \phi(\mathbf{h}_{q_i}, \mathbf{h}_{t_j^+}) > \phi(\mathbf{h}_{q_i}, \mathbf{h}_{t_i^+}) + \delta \;\text{or}\; j=i,\\
1, & \text{otherwise},
\end{cases}
\end{equation}
which yields the masked objective
\begin{equation}
\mathcal{L}_{\text{MedEmb}} \;=\; -\frac{1}{B}\sum_{i=1}^{B} \log
    \frac{\phi(\mathbf{h}_{q_i}, \mathbf{h}_{t_i^+})}
         {\phi(\mathbf{h}_{q_i}, \mathbf{h}_{t_i^+}) + \sum_{j=1}^{B} m_{ij}\,\phi(\mathbf{h}_{q_i}, \mathbf{h}_{t_j^+})}.
\end{equation}
Setting $m_{ij}=0$ on candidates that exceed the ground-truth positive by margin $\delta$ removes likely false negatives from the denominator while keeping all genuine negatives.

\paragraph{Dynamic batching constraint.}
Let $\ell_i^\star=\max(|\mathbf{q}_i|,|\mathbf{t}_i^+|)$ be the effective padded length of pair $i$. A batch $\mathcal{B}$ is accepted only if
\begin{equation}
2|\mathcal{B}|\cdot \max_{i\in\mathcal{B}}\ell_i^\star \le T,
\qquad
|\mathcal{B}|\le S_{\max},
\end{equation}
where the factor of $2$ accounts for query and target sequences and $T=2B_{\text{nom}}L_{\max}$ matches the token budget of static batching at nominal batch size $B_{\text{nom}}$ and maximum sequence length $L_{\max}$. Batches are formed online by greedy best-fit packing.

\FloatBarrier
\subsection{Family-level Dataset Details}\label{app:family-details}

This section expands the family-level summaries in Section~\ref{sec:tasks} with the full set of source datasets per retrieval family. Counts in parentheses give the total number of tasks and the modality breakdown (2D, 3D, TXT). The canonical dataset-to-task mapping is given in Appendix~\ref{app:datasets}.

\paragraph{Image--image retrieval (20 tasks).}
A 2D image or 3D volume query is ranked against studies from the same source, where a candidate is treated as positive if it shares the query's source-provided class, diagnosis, or anatomical label. The 2D subset draws from Kvasir~\cite{pogorelov2017kvasir}, LC25000~\cite{borkowski2019lung}, RFMiD~\cite{pachade2021retinal}, APTOS~\cite{aptos2019-blindness-detection}, MIMIC-CXR-LT~\cite{holste2022long}, ISIC-2019~\cite{codella2018skin,tschandl2018ham10000,hernandez2024bcn20000}, Brain-Tumor-MRI, PanNuke~\cite{gamper2019pannuke}, and the MedMNIST 2D collection (BloodMNIST, ChestMNIST, DermaMNIST, OCTMNIST, OrganAMNIST, OrganCMNIST, OrganSMNIST, PathMNIST, TissueMNIST)~\cite{yang2023medmnist}. The 3D subset uses CirrMRI600, MRNet~\cite{bien2018deep}, and Organ3dMNIST~\cite{yang2023medmnist}.

\paragraph{Modality-bridging visual retrieval (10 tasks).}
A query from one imaging modality is ranked against candidates from another, with positives defined by paired cross-modal visual matches. Pairs are constructed from same-patient or same-study annotations and cover three settings: MRI T1$\leftrightarrow$T2 sequence pairs from BraTS~\cite{kazerooni2024brain} and CirrMRI600~\cite{jha2025large}; MRI$\leftrightarrow$CT pairs from HaN-Seg~\cite{podobnik2023han} and SynthRAD2023 brain and pelvis~\cite{thummerer2023synthrad2023}; and 2D X-ray$\leftrightarrow$3D CT pairs derived from CT-RATE~\cite{hamamci2026generalist}.
Identifiers are used only to construct positives and are never provided to the model, so this family evaluates cross-modal visual matching rather than identifier lookup.

\paragraph{Image-to-label retrieval (22 tasks).}
An image or volume query is ranked against textual class labels. The 2D subset reuses most Image--image 2D sources (APTOS, BloodMNIST, Brain-Tumor-MRI, ChestMNIST, DermaMNIST, ISIC-2019, Kvasir, LC25000, MIMIC-CXR-LT, OCTMNIST, OrganAMNIST/C/S, PathMNIST, TissueMNIST) and adds CheXpert-Plus~\cite{chambon2024chexpert}. The 3D subset uses CT-RATE~\cite{hamamci2026generalist}, CirrMRI600, MRNet~\cite{bien2018deep}, NoduleMNIST, Organ3dMNIST, and SynapseMNIST~\cite{yang2023medmnist}.

\paragraph{Image-to-report retrieval (7 tasks).}
An image or volume query is ranked against free-form reports or captions. Sources span chest X-ray (MIMIC-CXR-report~\cite{johnson2019mimic}, CheXpert-Plus reports~\cite{chambon2024chexpert}), ultrasound (USData~\cite{li2024ultrasound}), 3D CT (M3D~\cite{bai2024m3d}, RadGen-CT~\cite{zhang2024radgenome}), 3D MRI (BraTS-MEN report annotations~\cite{labella2023asnr,lei2024autorg}), and dental imaging (MMDental~\cite{wang2025mmdental}).

\paragraph{Text-to-image retrieval (23 tasks).}
A class label or short clinical description is ranked against image candidates from the same dataset, evaluating the reverse direction of image--text alignment. Tasks reuse the image-side sources from Image-to-label and Image-to-report (APTOS, BloodMNIST, Brain-Tumor-MRI, ChestMNIST, CheXpert-Plus, DermaMNIST, ISIC-2019, Kvasir, LC25000, MIMIC-CXR-LT, OCTMNIST, OrganAMNIST/C/S, PanNuke, PathMNIST, Retinal/RFMiD, TissueMNIST), together with the volumetric sources CT-RATE, CirrMRI600, M3D, MRNet, and Organ3dMNIST.

\paragraph{Question-conditioned retrieval (11 tasks).}
Medical VQA is recast as answer retrieval, where an image--question pair queries the dataset's unique answer set. The 2D subset spans MIMIC-CXR-VQA~\cite{PhysioNet-medical-cxr-vqa-dataset-1.0.0}, MedPIX~\cite{siragusa2025medpix}, OmniMedVQA~\cite{hu2024omnimedvqa}, PMC-VQA~\cite{zhang2023pmc}, Path-VQA~\cite{he2020pathvqa}, ROCO-VQA~\cite{pelka2018roco}, RadImageNet-VQA~\cite{butsanets2025radimagenet}, and VQA-RAD~\cite{lau2018dataset}. The 3D subset adds CT-RATE-VQA~\cite{hamamci2026generalist}, M3D-3DQA~\cite{bai2024m3d}, and RadGen-CT VQA~\cite{zhang2024radgenome}.

\paragraph{Region-level text-to-region retrieval (7 tasks).}
Visual grounding is reformulated as ranking source-provided region crops against a query composed of the full image and a region-level text description. Sources cover chest X-ray (Chest-ImaGenome~\cite{PhysioNet-chest-imagenome-1.0.0}, VinDr-CXR~\cite{PhysioNet-vindr-cxr-1.0.0}), endoscopy (Kvasir-SEG~\cite{jha2019kvasir}), pathology (PanNuke region annotations~\cite{gamper2019pannuke}), dermatology (ISIC-2018~\cite{codella2019skin,tschandl2018ham10000}), ultrasound (ultrasound nerve segmentation~\cite{ultrasound-nerve-segmentation}), and mammography (VinDr-Mammo~\cite{PhysioNet-vindr-mammo-1.0.0}); region crops are standardized through MedSegBench~\cite{kucs2024medsegbench}.

\paragraph{Biomedical text--text retrieval (7 tasks).}
Query and candidates are biomedical text passages from the same source, covering medical QA, radiology findings-to-impression retrieval, and clinical knowledge retrieval. Sources are MedMCQA~\cite{pal2022medmcqa}, PubMedQA~\cite{jin2019pubmedqa}, MedicalQARetrieval, PublicHealthQA, MIMIC-CXR findings-to-impression~\cite{johnson2019mimic}, CheXpert-Plus findings-to-impression~\cite{chambon2024chexpert}, and MMDental text--text~\cite{wang2025mmdental}.

\FloatBarrier
\subsection{Cross-Family Consistency}\label{app:consistency}

To quantify whether a model is balanced across retrieval families rather than only achieving a high average, we report the coefficient of variation
\[
\mathrm{CV}(m) \;=\; \frac{\sigma_m}{\mu_m + \epsilon},
\]
where $\mu_m$ and $\sigma_m$ are the mean and standard deviation of model $m$'s Hit@1 scores across the eight retrieval families (I--I, SP--CM, I--Lab, I--Rep, T--I, VQA, VG, T--T). Lower CV indicates a more consistent model. Table~\ref{tab:cv} reports $\mu$, $\sigma$, CV, and the per-family Hit@1 scores used to compute them.

\begin{table}[h]
\centering
\caption{Cross-family consistency statistics. $\mu$ and $\sigma$ are computed from family-level Hit@1 across the eight retrieval families; lower CV $= \sigma/\mu$ indicates more balanced performance.}
\label{tab:cv}
\resizebox{\textwidth}{!}{
\begin{tabular}{lccc cccccccc}
\toprule
Model & $\mu$ & $\sigma$ & CV & I-I & MB-VR & I-Lab & I-Rep & T-I & VQA & VG & T-T \\
\midrule
\multicolumn{12}{c}{\textbf{Baseline Models}} \\
\midrule
siglip-so400m-patch14-384  & 11.34 &  9.45 & 0.833 & 31.38 &  6.78 & 12.19 &  5.01 & 20.91 &  2.36 &  2.65 &  9.45 \\
blip2-itm-vit-g            &  9.54 &  8.16 & 0.855 & 29.67 &  7.23 &  7.05 &  5.65 & 12.85 &  3.24 &  2.70 &  7.91 \\
MedImageInsight            & 18.68 & 12.12 & 0.649 & 44.74 & 10.71 & 14.25 &  9.90 & 26.13 &  9.81 &  7.05 & 26.84 \\
MedSigLIP-448              & 10.55 &  7.72 & 0.732 & 19.73 &  3.72 & 13.51 &  4.86 & 24.18 &  6.49 &  0.46 & 11.41 \\
RzenEmbed                  & 18.71 & 16.93 & 0.905 & 31.04 &  8.62 & 13.86 &  7.99 & 17.25 &  4.49 &  7.96 & 58.47 \\
VLM2Vec-V2.0               & 16.24 & 11.14 & 0.686 & 26.27 &  6.18 & 14.33 &  7.22 & 15.60 & 16.80 &  3.84 & 39.71 \\
Qwen3-VL-Embed-2B      & 21.91 & 15.74 & 0.718 & 33.04 & 10.60 & 18.90 &  9.82 & 21.45 & 20.08 &  4.10 & 57.32 \\
Qwen3-VL-Embed-8B      & 25.31 & 16.56 & 0.654 & 34.69 &  8.99 & 20.69 & 11.53 & 20.09 & 23.07 & 18.71 & 64.72 \\
\midrule
\multicolumn{12}{c}{\textbf{MedEmb}} \\
\midrule
MedEmb-2B                  & 41.63 & 16.50 & 0.396 & 52.03 & 14.49 & 56.43 & 25.76 & 52.75 & 32.28 & 33.47 & 65.81 \\
MedEmb-8B                  & 46.06 & 15.81 & 0.343 & 54.36 & 20.30 & 58.83 & 28.97 & 55.70 & 35.47 & 44.17 & 70.64 \\
\bottomrule
\end{tabular}
}
\end{table}

\FloatBarrier
\subsection{Hit@5 Results}\label{app:hit5}
For completeness we report Hit@5 per retrieval family and across the ID, held-out-source, and All splits in Table~\ref{tab:hit5}. The trends mirror the Hit@1 / MRR@5 results in Table~\ref{tab:main_results_medical}: MedEmb-8B is the strongest model on every family and on every split.

\begin{table}[h]
\centering
\caption{Task-macro Hit@5 (\%) on MedHEB. Family and summary columns follow Table~\ref{tab:main_results_medical}, providing a top-5 complement to Hit@1 and MRR@5.}
\label{tab:hit5}
\resizebox{\textwidth}{!}{
\begin{tabular}{lcccccccccccc}
\toprule
\textbf{Model}
& \multicolumn{8}{c}{\textbf{Retrieval Families}}
& \multicolumn{3}{c}{\textbf{Summary}} \\
\cmidrule(r){2-9} \cmidrule(r){10-12}
& I-I & MB-VR & I-Lab & I-Rep & T-I & VQA & VG & T-T
& ID & \makecell{Held-out-\\source} & All \\
\midrule
\textbf{\# of Tasks $\rightarrow$}
& 20 & 10 & 22 & 7 & 23 & 11 & 7 & 7
& 96 & 11 & 107 \\
\midrule
\multicolumn{12}{c}{\textbf{Baseline Models}} \\
\midrule
SigLIP~\cite{zhai2023sigmoid}
& 77.43 & 20.22 & 42.01 & 14.10 & 64.23 &  9.23 &  7.00 & 21.74
& 41.61 & 50.83 & 42.56 \\

BLIP2~\cite{li2023blip}
& 77.53 & 21.65 & 33.17 & 22.59 & 57.28 &  9.22 &  6.76 & 18.29
& 38.33 & 51.75 & 39.71 \\

MedImageInsight~\cite{codella2024medimageinsight}
& 84.23 & 33.63 & 42.95 & 31.03 & 67.58 & 20.51 & 12.58 & 47.39
& 48.68 & 64.48 & 50.31 \\

MedSigLIP~\cite{sellergren2025medgemma}
& 70.11 & 14.78 & 41.66 & 18.89 & 66.98 & 15.08 &  2.27 & 27.19
& 40.59 & 55.84 & 42.16 \\

RzenEmbed~\cite{jian2025rzenembed}
& 77.64 & 25.72 & 43.14 & 30.44 & 60.11 & 14.01 & 13.99 & 72.62
& 46.77 & 56.83 & 47.80 \\

VLM2Vec-V2.0~\cite{meng2025vlm2vec}
& 73.88 & 25.52 & 49.09 & 27.89 & 59.66 & 31.78 &  8.45 & 58.39
& 47.85 & 54.94 & 48.58 \\

Qwen3-VL-Embed-2B~\cite{li2026qwen3}
& 78.73 & 30.91 & 54.75 & 33.52 & 63.00 & 40.43 &  8.00 & 70.26
& 53.32 & 58.69 & 53.87 \\

Qwen3-VL-Embed-8B~\cite{li2026qwen3}
& 79.62 & 30.08 & 56.59 & 32.69 & 62.67 & 46.70 & 28.07 & 78.05
& 56.10 & 61.76 & 56.68 \\

\midrule
\multicolumn{12}{c}{\textbf{MedEmb}} \\
\midrule

MedEmb-2B
& 86.72 & 44.98 & 91.91 & 53.96 & 81.18 & 59.33 & 49.79 & 82.35
& 75.38 & 71.99 & 75.03 \\

MedEmb-8B
& \textbf{87.92} & \textbf{50.52} & \textbf{92.48} & \textbf{62.00} & \textbf{83.75} & \textbf{64.04} & \textbf{60.89} & \textbf{87.14}
& \textbf{79.16} & \textbf{72.74} & \textbf{78.50} \\
\bottomrule
\end{tabular}
}
\end{table}

\FloatBarrier
\subsection{Text Granularity and Embedding Compositionality}\label{app:granularity}
\begin{figure}[h]
  \centering
  \includegraphics[width=\linewidth]{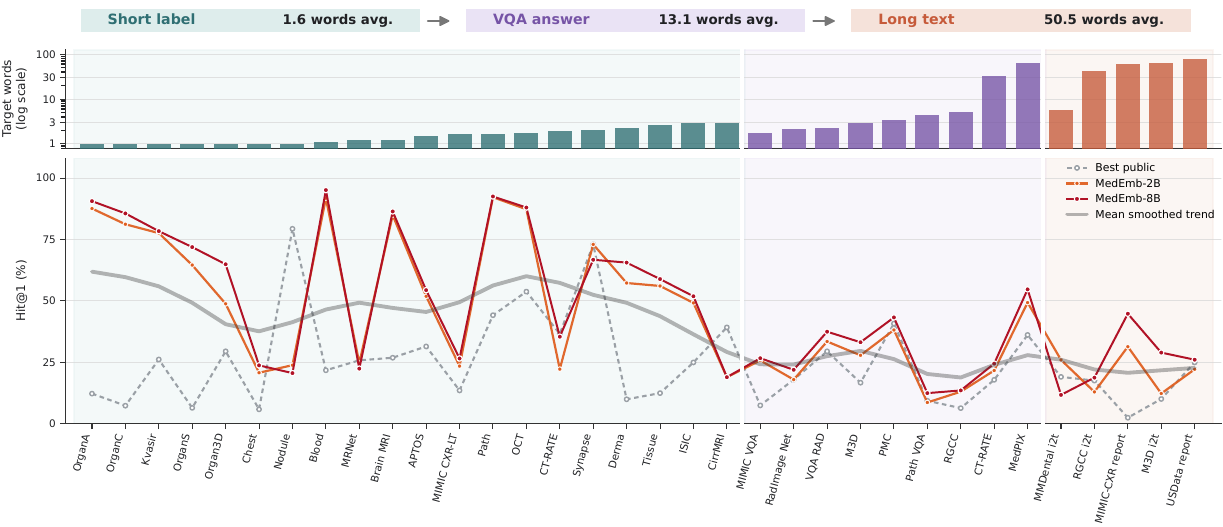}
  \caption{Target granularity effects in text-conditioned retrieval. Tasks are ordered by average target length and grouped into short labels, VQA answers, and long-text targets; lower panels compare Hit@1 trajectories for the best public baseline, MedEmb-2B, and MedEmb-8B.}
  \label{fig:granularity}
\end{figure}

Target format exposes another embedding bottleneck: the ability to align visual evidence with increasingly compositional clinical language. We stratify text-conditioned tasks into short labels, question-conditioned answers, and long-text targets (Figure~\ref{fig:granularity}). Short labels usually encode one dominant clinical concept, making nearest-neighbor alignment relatively direct. Question-conditioned answers add attributes but constrain relevance through the question. Long-text targets, such as reports or descriptions, bundle multiple clinical concepts at once, including anatomy, morphology, severity, laterality, uncertainty, and co-occurring findings. As target length and specificity increase, task-level Hit@1 decreases and becomes more variable across models. This pattern suggests that difficulty grows not merely because the target text is longer, but because the embedding must compress several clinically relevant attributes into a single vector while preserving their correspondence to the same visual evidence. Long-text retrieval therefore highlights a remaining bottleneck for medical embeddings: compositional alignment between rare, localized, and subtle visual findings and precise clinical language.

\FloatBarrier
\subsection{Per-task Instruction Templates}\label{app:instructions}
Each MedHEB task is encoded with two strings: \texttt{qry\_inst}, the instruction prepended to the query, and \texttt{qry\_text}, the per-task or per-row content. Within each retrieval family, instructions are largely shared; we list the shared 2D instructions below and tabulate the per-task overrides for image classification, 3D volume tasks, and biomedical text--text tasks in Tables~\ref{tab:inst-cls}, \ref{tab:inst-3d}, and \ref{tab:inst-text}.

\textbf{Image--image (same-category)} --- MIMIC-CXR-LT, ISIC-2019, Brain-Tumor-MRI, the medical-MNIST family, Kvasir, PanNuke, APTOS, RFMiD, LC25000: \texttt{qry\_inst} = ``\textit{Find the same category image with the provided image:}'', \texttt{qry\_text} = \texttt{<image>} (the image itself serves as the query content).

\textbf{Text-to-image} (all 2D T2I tasks): \texttt{qry\_inst} = ``\textit{Find an image that matches the given category:}'', \texttt{qry\_text} = category label.

\textbf{Image classification (image-to-label)}: \texttt{qry\_inst} = ``\textit{Represent the given image with the following question:}'', with the task-specific \texttt{qry\_text} listed in Table~\ref{tab:inst-cls}.

\textbf{Visual question answering} --- Path-VQA, PMC-VQA, ROCO-VQA, MedPIX, RadImageNet-VQA, VQA-RAD, MIMIC-CXR-VQA, OmniMedVQA: \texttt{qry\_inst} = ``\textit{Represent the given image with the following question:}'', \texttt{qry\_text} = per-row question.

\textbf{Visual grounding} --- PanNuke-VG, UltrasoundNerve, Chest-ImaGenome, Gastrointestinal, SkinLesion, VinDr-CXR, VinDr-Mammo: \texttt{qry\_inst} = ``\textit{Locate the specific region that corresponds to the provided text description.}'', \texttt{qry\_text} = category or anatomical descriptor.

\textbf{Image-to-report} --- MIMIC-CXR-report and CheXpert-Plus i2t share ``\textit{Represent the given chest X-ray image for radiology report retrieval:}''; USData uses ``\textit{Represent the given ultrasound image list for radiology report retrieval:}''.

\textbf{Cross-source graded retrieval} (MIMIC-CXR\,$\to$\,CheXpert-Plus, \S\ref{sec:graded-ood}): \texttt{qry\_inst} = ``\textit{Find chest X-ray images that share the most disease findings with the given image:}'', \texttt{qry\_text} = \texttt{<image>} (the image itself serves as the query content).

\begin{table}[h]
\centering
\small
\caption{2D image-to-label query texts. All rows share \texttt{qry\_inst} = ``\textit{Represent the given image with the following question:}''; the table lists task-specific \texttt{qry\_text} prompts.}
\label{tab:inst-cls}
\begin{tabular}{p{4.5cm} p{8.5cm}}
\toprule
Task(s) & \texttt{qry\_text} \\
\midrule
MIMIC-CXR-LT, ChestMNIST, CheXpert-Plus & What is the radiological abnormal findings observed in this chest X-ray? \\
ISIC-2019 & What is the dermatological condition observed in this image? \\
Brain-Tumor-MRI & What is the brain tumor diagnostic category observed in this MRI scan? \\
BloodMNIST & What is the blood cell category observed in this image? \\
DermaMNIST & What is the dermatological diagnostic category observed in this image? \\
OCTMNIST & What is the retinal diagnostic category observed in this image? \\
OrganAMNIST / OrganCMNIST / OrganSMNIST & What is the anatomical structure observed in this image? \\
PathMNIST & What is the histopathological tissue category observed in this image? \\
TissueMNIST & What is the renal cellular category observed in this image? \\
Kvasir & What is the endoscopic finding observed in this image? \\
APTOS & What is the diabetic retinopathy severity level observed in this retinal image? \\
LC25000 & What is the diagnostic category observed in this histopathological image? \\
\bottomrule
\end{tabular}
\end{table}

\begin{table}[h]
\centering
\small
\caption{3D instruction templates by retrieval family. ``per row'' denotes instance-specific question or report text; \texttt{<image>} denotes the sampled 3D volume or image sequence used as query content.}
\label{tab:inst-3d}
\setlength{\tabcolsep}{3pt}
\renewcommand{\arraystretch}{1.12}
\begin{tabularx}{\textwidth}{@{}>{\raggedright\arraybackslash}p{3.35cm} >{\raggedright\arraybackslash}X >{\raggedright\arraybackslash}p{2.85cm}@{}}
\toprule
Task(s) & \texttt{qry\_inst} & \texttt{qry\_text} \\
\midrule
\multicolumn{3}{l}{\textit{Volume classification}} \\
CT-RATE-cls & Represent the given list of images with the following question: & What is the pathological finding represented in this list of images? \\
CirrMRI600-cls & Represent the given MRI series with the following question: & What is the degree of liver cirrhosis shown in the picture? \\
MRNet-cls & Represent the given MRI series with the following question: & What type of knee disease is shown in this image? \\
NoduleMNIST-cls & Represent the given CT series with the following question: & What is the nature of the tumor in the chest? \\
Organ3dMNIST-cls & Represent the given CT series with the following question: & What anatomical structure is shown in this image? \\
SynapseMNIST-cls & Represent the given Electron Microscope with the following question: & What type of synapse is this? \\
\midrule
\multicolumn{3}{l}{\textit{Volume question answering}} \\
CT-RATE-vqa, RGCC-vqa, M3D-3dqa & Represent the given CT series with the following question: & per row \\
\midrule
\multicolumn{3}{l}{\textit{Volume-to-report}} \\
RGCC-i2t, M3D-i2t & Find a detailed report summarizing the given CT series: & \texttt{<image>} \\
MMDental-i2t & Find a detailed diagnosis summarizing the given CBCT series: & \texttt{<image>} \\
BraTS-MEN-i2t & Find a detailed report summarizing the given Brain MRI series: & \texttt{<image>} \\
\midrule
\multicolumn{3}{l}{\textit{Text-to-volume}} \\
CT-RATE-t2i, M3D-t2i & Find a CT series that matches the given report: & per row \\
CirrMRI600-t2i, MRNet-t2i & Find an MRI series that matches the given category: & category \\
Organ3dMNIST-t2i & Find an CT series that matches the given category: & category \\
\midrule
\multicolumn{3}{l}{\textit{Volume--volume}} \\
CT-RATE-i2i & Find a CT series that matches the given chest X-ray image: & \texttt{<image>} \\
CirrMRI600-i2i, MRNet-i2i & Find the same category MRI series with the provided MRI series: & \texttt{<image>} \\
Organ3dMNIST-i2i & Find the same category CT series with the provided CT series: & \texttt{<image>} \\
BraTS-t1$\to$t2, CirrMRI600-t1$\to$t2 & Find a T2-weighted MRI series that matches the given T1-weighted MRI series: & \texttt{<image>} \\
BraTS-t2$\to$t1, CirrMRI600-t2$\to$t1 & Find a T1-weighted MRI series that matches the given T2-weighted MRI series: & \texttt{<image>} \\
HaN-Seg mri$\to$ct, SynthRAD mri2ct & Find a CT series that matches the given MRI series: & \texttt{<image>} \\
SynthRAD ct2mri & Find an MRI series that matches the given CT series: & \texttt{<image>} \\
\bottomrule
\end{tabularx}
\end{table}

\begin{table}[ht]
\centering
\small
\caption{Biomedical text--text instruction templates. Rows list the shared \texttt{qry\_inst} for each source; \texttt{qry\_text} is populated per instance with the corresponding question, finding, context, or diagnosis.}
\label{tab:inst-text}
\begin{tabular}{p{4cm} p{8cm}}
\toprule
Task(s) & \texttt{qry\_inst} \\
\midrule
MedMCQA, MedicalQA, PublicHealthQA & Find the answer of the given question: \\
MIMIC findings-to-impression, CheXpert-Plus f2i & Identifying impressions based on given radiology findings. \\
PubMedQA & Summarize the content of the given context: \\
MMDental & Find the treatment plan of the given diagnosis: \\
\bottomrule
\end{tabular}
\end{table}

\FloatBarrier
\subsection{Dataset Details}\label{app:datasets}
\begin{figure}[h]
  \centering
  \includegraphics[width=\linewidth]{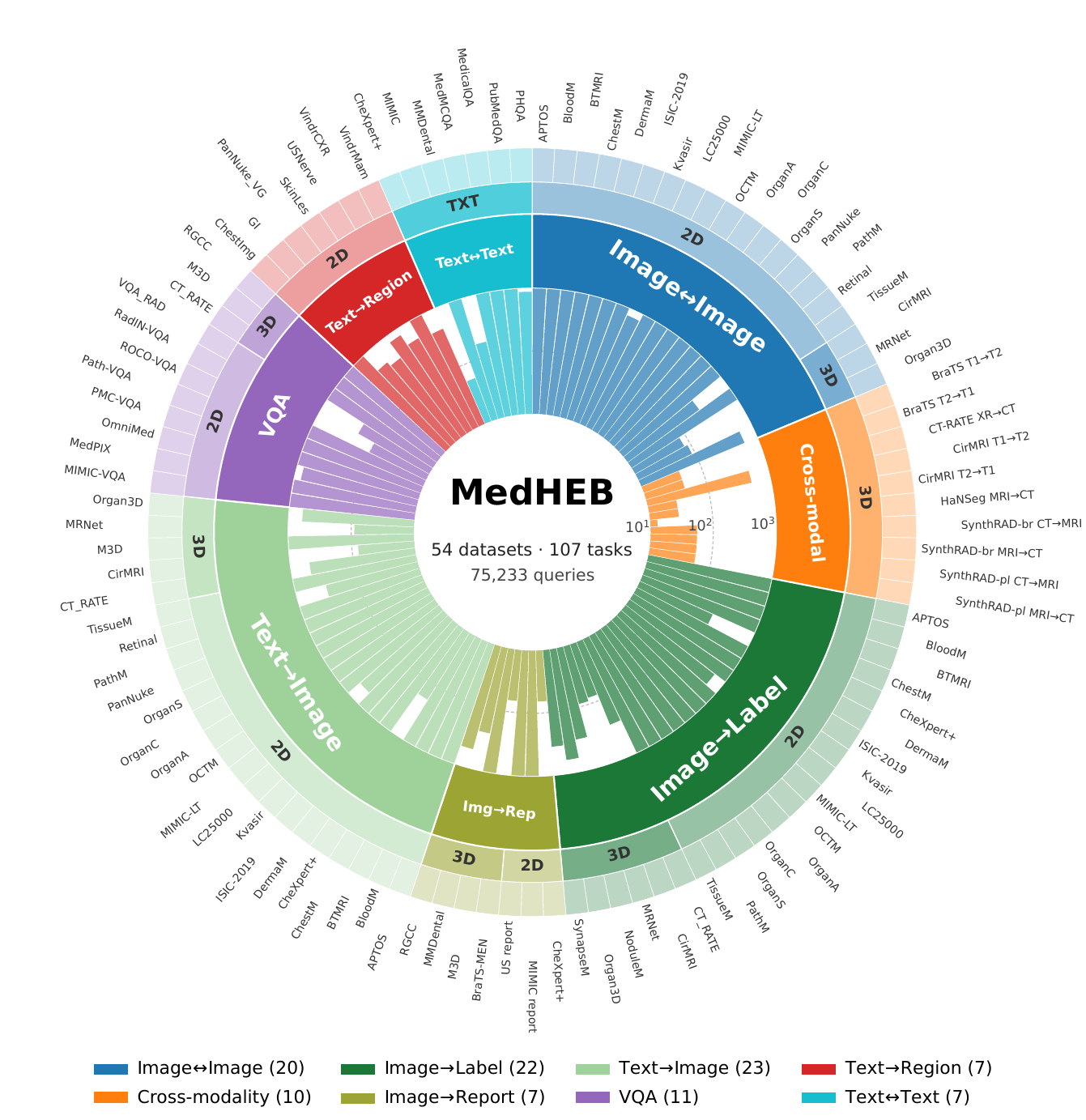}
  \caption{MedHEB benchmark composition.}
  \label{fig:app_composition}
\end{figure}

This section lists the source datasets that compose MedHEB, including the dataset name, evidence form (2D, 3D, or text), modality, and a publicly accessible URL. MedHEB is built from \textbf{54 source datasets} spanning 2D images, 3D volumes, and biomedical text. Figure~\ref{fig:app_composition} summarizes the composition of MedHEB.

{\renewcommand{\arraystretch}{1.15}
\footnotesize
\begin{xltabular}{\textwidth}{@{}p{2.6cm} p{2.4cm} p{3.6cm} >{\raggedright\arraybackslash}X@{}}
\caption{Source datasets composing MedHEB. Datasets are grouped by evidence form; modality, license / use terms, and official access URL specify the reconstruction manifest used for benchmark release. Where the licensing terms are in conflict across sources or otherwise unclear, the entry is marked N/A and the access URL remains authoritative.}\label{tab:dataset_details}\\
\toprule
\textbf{Dataset} & \textbf{Modality} & \textbf{License / use terms} & \textbf{Link} \\
\midrule
\endfirsthead
\multicolumn{4}{@{}l@{}}{\textit{(continued from previous page)}} \\
\toprule
\textbf{Dataset} & \textbf{Modality} & \textbf{License / use terms} & \textbf{Link} \\
\midrule
\endhead
\midrule
\multicolumn{4}{@{}r@{}}{\textit{(continued on next page)}} \\
\endfoot
\bottomrule
\endlastfoot
\multicolumn{4}{@{}l@{}}{\textit{2D images}} \\
\midrule
APTOS-2019              & Fundus photography           & Competition Use and Non-Commercial \& Academic Research & \url{https://www.kaggle.com/competitions/aptos2019-blindness-detection/overview} \\
BloodMNIST              & Blood cell microscopy        & CC BY 4.0                          & \url{https://zenodo.org/records/10519652} \\
Brain-Tumor-MRI         & Brain MRI                    & CC BY 4.0                          & \url{https://www.kaggle.com/datasets/masoudnickparvar/brain-tumor-mri-dataset} \\
Chest-ImaGenome         & Chest X-ray                  & PhysioNet Credentialed Health Data License 1.5.0 & \url{https://physionet.org/content/chest-imagenome/1.0.0/} \\
ChestMNIST              & Chest X-ray                  & CC BY 4.0                          & \url{https://zenodo.org/records/10519652} \\
CheXpert-Plus           & Chest X-ray                  & Stanford Dataset Research Use Agreement & \url{https://stanfordaimi.azurewebsites.net/datasets/5158c524-d3ab-4e02-96e9-6ee9efc110a1} \\
DermaMNIST              & Skin disease image           & CC BY-NC 4.0                       & \url{https://zenodo.org/records/10519652} \\
ISIC-2019               & Skin disease image           & CC BY-NC 4.0                       & \url{https://challenge.isic-archive.com/data/} \\
Isic2018MSBench         & Skin image                   & CC BY-NC 4.0                       & \url{https://github.com/zekikus/MedSegBench} \\
Kvasir                  & GI tract endoscopy           & Simula custom (research / educational; commercial use requires written permission) & \url{https://datasets.simula.no/kvasir/} \\
Kvasir-SEG              & GI tract endoscopy           & Simula custom (research / educational; commercial use requires written permission) & \url{https://datasets.simula.no/kvasir-seg/} \\
LC25000                 & Pathology                    & N/A                                & \url{https://www.kaggle.com/datasets/javaidahmadwani/lc25000} \\
MedPIX-VQA              & Multi-modality VQA           & Apache 2.0                         & \url{https://huggingface.co/datasets/adishourya/MEDPIX-ClinQA} \\
MIMIC-CXR               & Chest X-ray                  & PhysioNet Credentialed Health Data License 1.5.0 & \url{https://physionet.org/content/mimic-cxr/2.1.0/} \\
MIMIC-CXR-LT            & Chest X-ray                  & PhysioNet Credentialed Health Data License 1.5.0 & \url{https://physionet.org/content/mimic-cxr/2.1.0/} \\
MIMIC-CXR-VQA           & Chest X-ray VQA              & PhysioNet Credentialed Health Data License 1.5.0 (per underlying MIMIC-CXR) & \url{https://physionet.org/content/medical-cxr-vqa-dataset/1.0.0/} \\
OCTMNIST                & Optical coherence tomography & CC BY 4.0                          & \url{https://zenodo.org/records/10519652} \\
OmniMedVQA              & Multi-modality VQA           & Per source datasets (no single license) & \url{https://huggingface.co/datasets/foreverbeliever/OmniMedVQA} \\
OrganAMNIST (LiTS)      & Abdominal CT (axial)         & CC BY 4.0                          & \url{https://zenodo.org/records/10519652} \\
OrganCMNIST (LiTS)      & Abdominal CT (coronal)       & CC BY 4.0                          & \url{https://zenodo.org/records/10519652} \\
OrganSMNIST (LiTS)      & Abdominal CT (sagittal)      & CC BY 4.0                          & \url{https://zenodo.org/records/10519652} \\
PanNuke                 & Pathology                    & CC BY-NC-SA 4.0                    & \url{https://huggingface.co/datasets/RationAI/PanNuke} \\
Path-VQA                & Pathology VQA                & MIT                                & \url{https://huggingface.co/datasets/flaviagiammarino/path-vqa} \\
PathMNIST               & Colon pathology              & CC BY 4.0                          & \url{https://zenodo.org/records/10519652} \\
PMC-VQA                 & Multi-modality VQA           & CC BY-SA                                & \url{https://huggingface.co/datasets/RadGenome/PMC-VQA} \\
PolypGenMSBench         & GI tract endoscopy           & CC BY-NC 4.0                       & \url{https://github.com/zekikus/MedSegBench} \\
RadImageNet-VQA         & Multi-modality VQA           & Apache-2.0
 & \url{https://huggingface.co/datasets/raidium/RadImageNet-VQA} \\
RFMiD                   & Fundus photography           & Other                         & \url{https://www.kaggle.com/datasets/andrewmvd/retinal-disease-classification} \\
ROCO-VQA                & Multi-modality VQA           & Apache 2.0                         & \url{https://huggingface.co/datasets/adishourya/ROCO-QA} \\
TissueMNIST             & Kidney cortex microscopy     & CC BY 4.0                          & \url{https://zenodo.org/records/10519652} \\
UltrasoundNerveMSBench  & Ultrasound                   & CC BY-NC 4.0                     & \url{https://github.com/zekikus/MedSegBench} \\
USData                  & Ultrasound                   & N/A                                & \url{https://github.com/LijunRio/Ultrasound-Report-Generation} \\
USforKidneyMSBench      & Ultrasound                   & CC BY-NC 4.0                       & \url{https://github.com/zekikus/MedSegBench} \\
UWSkinCancerMSBench     & Skin image                   & CC BY-NC 4.0                       & \url{https://github.com/zekikus/MedSegBench} \\
VinDr-CXR               & Chest X-ray                  & PhysioNet Credentialed Health Data License 1.5.0 & \url{https://physionet.org/content/vindr-cxr/1.0.0/} \\
VinDr-Mammo             & Breast mammography           & PhysioNet Credentialed Health Data License 1.5.0 & \url{https://physionet.org/content/vindr-mammo/1.0.0/} \\
VQA-RAD                 & Multi-modality VQA           & CC0 1.0                            & \url{https://huggingface.co/datasets/flaviagiammarino/vqa-rad} \\
\midrule
\multicolumn{4}{@{}l@{}}{\textit{3D volumes}} \\
\midrule
BraTS-MEN               & Brain MRI                    & CC BY-NC                           & \url{https://www.synapse.org/Synapse:syn51514106} \\
BraTS-PED               & Brain MRI                    & CC-BY-NC                                & \url{https://www.synapse.org/Synapse:syn51514108} \\
CirrMRI600              & Liver MRI                    & Other                                & \url{https://osf.io/cuk24/overview} \\
CT-RATE                 & Chest CT                     & CC BY-NC-SA 4.0 (gated; no commercial; no redistribution) & \url{https://huggingface.co/datasets/ibrahimhamamci/CT-RATE} \\
HaN-Seg                 & Head \& neck MRI/CT          & CC BY-NC-ND 4.0                    & \url{https://zenodo.org/records/7442914} \\
M3D                     & Multi-modality 3D            & Apache License 2.0                                & \url{https://www.modelscope.cn/datasets/GoodBaiBai88/M3D-Cap} \\
MMDental                & Dental CBCT                  & CC BY 4.0      & \url{https://springernature.figshare.com/articles/dataset/28505276} \\
MRNet                   & Knee MRI                     & Stanford Research Use Agreement (non-commercial research; no redistribution) & \url{https://aimi.stanford.edu/datasets/mrnet-knee-mris} \\
NoduleMNIST             & Chest CT                     & CC BY 4.0                          & \url{https://zenodo.org/records/10519652} \\
OrganMNIST3D (LiTS)     & Abdominal CT                 & CC BY 4.0                          & \url{https://zenodo.org/records/10519652} \\
RadGenome-Chest-CT      & Chest CT                     & CC BY 4.0  
 & \url{https://huggingface.co/datasets/RadGenome/RadGenome-ChestCT} \\
SynapseMNIST (MitoEM)   & Pathology (electron microscopy) & CC BY 4.0                       & \url{https://zenodo.org/records/10519652} \\
SynthRAD2023            & Brain \& pelvis MRI/CT       & CC BY-NC 4.0                       & \url{https://zenodo.org/records/7260705} \\
\midrule
\multicolumn{4}{@{}l@{}}{\textit{Biomedical text}} \\
\midrule
MedicalQA               & Biomedical text              & CC0 1.0                            & \url{https://huggingface.co/datasets/mteb/medical_qa} \\
MedMCQA                 & Biomedical text              & Apache 2.0                         & \url{https://huggingface.co/datasets/openlifescienceai/medmcqa} \\
PubMedQA                & Biomedical text              & MIT                                & \url{https://huggingface.co/datasets/qiaojin/PubMedQA} \\
PublicHealthQA          & Biomedical text              & CC BY 3.0 (CDC) / CC BY-NC-SA 3.0 (WHO) & \url{https://huggingface.co/datasets/xhluca/publichealth-qa} \\
\end{xltabular}
}

\FloatBarrier
\subsection{Reproducibility, Release Scope, and Ethics}\label{app:release}

\paragraph{Reproducibility and artifact scope.}
Sections~\ref{sec:construction}, \ref{sec:baseline}, \ref{sec:metrics}, and \ref{sec:main-results} specify the task construction protocol, train/validation/test split design, evaluated models, metrics, and MedEmb training recipe. The accompanying MedHEB artifact provides task indices, preprocessing scripts, evaluation code, and source-specific access manifests; because many component datasets are distributed under their own licenses, data-use agreements, or credentialing requirements, the benchmark release does not re-host restricted raw images, volumes, or reports.

\paragraph{Compute and statistical reporting.}
MedEmb-2B and MedEmb-8B are each trained for one epoch on the MedHEB ID split using LoRA adapters, AdamW with cosine decay and warmup, token-budgeted dynamic batching, and $8$ NVIDIA H100 GPUs (Section~\ref{sec:baseline}). Main benchmark metrics are deterministic evaluations over fixed query/candidate pools; for the clinician preference study in Section~\ref{sec:human-eval}, MedEmb-8B wins $260$ of $382$ non-tied comparisons, corresponding to a two-sided exact binomial test with $p<10^{-11}$.

\paragraph{Limitations, broader impacts, and safeguards.}
The benchmark improves coverage of heterogeneous clinical retrieval, but it remains limited by the quality, ontology design, demographic composition, and modality mix of its source datasets, and by the fact that clinically acceptable evidence can be broader than a single annotated positive. Positive impacts include more reproducible evaluation of multimodal medical embeddings and clearer diagnosis of where current retrieval models fail; negative impacts may arise if benchmark scores are interpreted as evidence of clinical readiness, if source-data biases are overlooked, or if retrieved neighbors are used without appropriate expert review. MedHEB therefore releases metadata and access manifests rather than restricted source data, and any downstream use should remain subject to the original asset terms and human clinical oversight.

\FloatBarrier
\subsection{Graded-Relevance Protocol}\label{app:graded-protocol}

The graded-relevance evaluation in Section~\ref{sec:graded-relevance} uses the following annotation-based relevance weights.

\textbf{Severity.} Relevance is $1$ for the same severity stage, $0.5$ for a one-stage difference, $0.25$ for a two-stage difference, and $0$ for larger differences.

\textbf{Laterality.} Relevance is $1$ for the same side, $0.5$ for the contralateral side of the same organ, and $0$ for a different organ.

\textbf{Diagnosis.} Relevance is $1$ for the same diagnosis, $0.5$ for a different diagnosis within the same source-defined category, and $0$ for unrelated diagnoses.

\textbf{Cell/tissue hierarchy.} Relevance is $1$ for an exact label match, $0.5$ for labels sharing the same published hierarchy, and $0$ otherwise.

\subsection{Human Evaluation Protocol}\label{app:human-protocol}

The clinician preference study in Section~\ref{sec:human-eval} uses $530$ stratified queries sampled across the $107$ MedHEB tasks. For each query, MedEmb-8B and Qwen3-VL-Embed-8B each produce a top-3 retrieval list. The two lists are presented side by side as ``List A'' and ``List B'' in a per-query randomized order, and model identity, gold answer, and task stratum are hidden from the evaluator. The evaluator selects ``List A is better'', ``List B is better'', or ``Tie'', and may add an optional free-text comment.

\begin{figure}[h]
  \centering
  \includegraphics[width=0.85\linewidth]{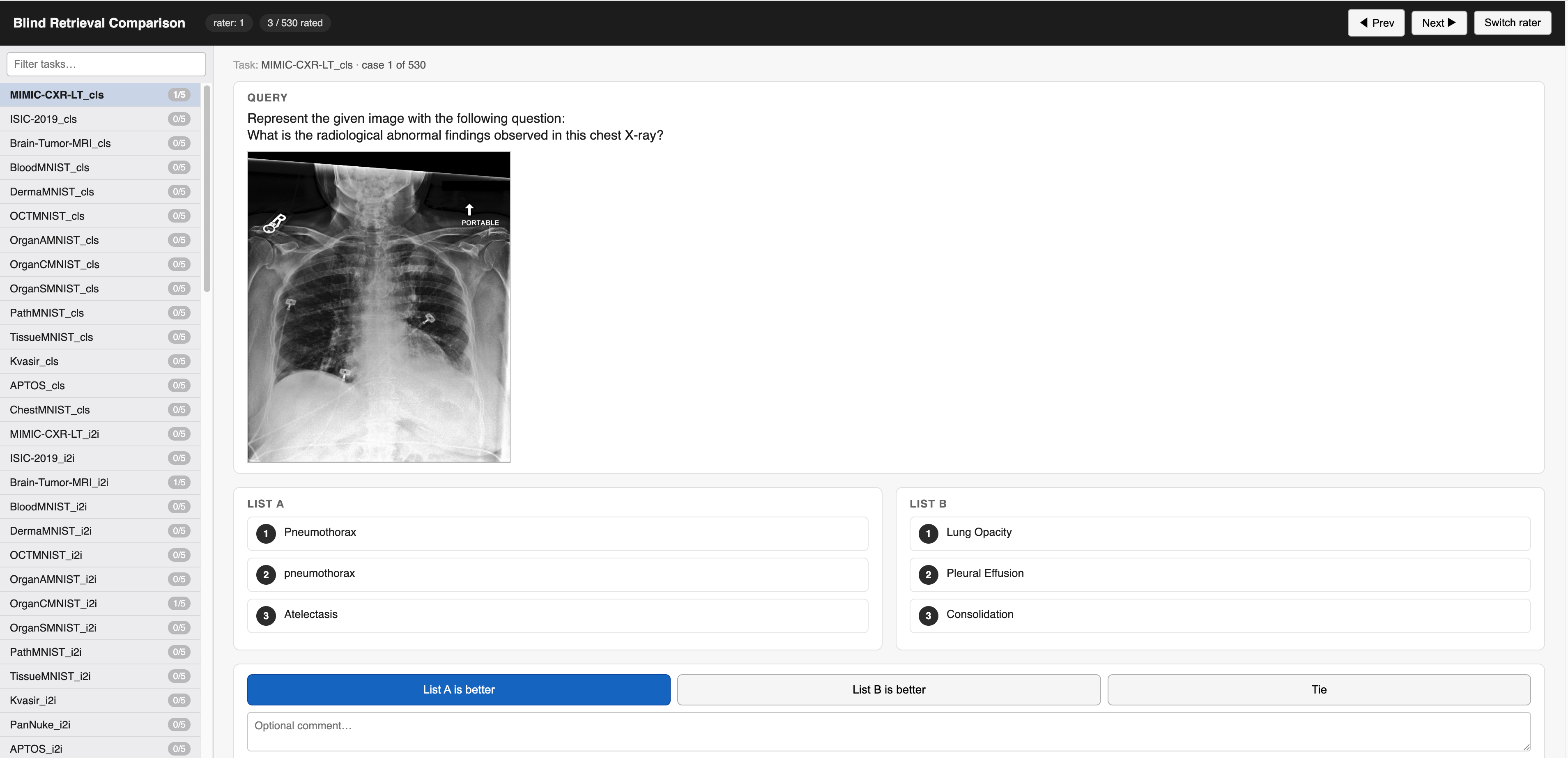}
  \caption{Blind-comparison interface for 2D queries. The evaluator sees the system prompt, the query image, two top-3 candidate lists shown in randomized A/B order, and three preference buttons (``List A is better'', ``List B is better'', ``Tie'') with an optional comment field. Model identity, gold answer, and task stratum are hidden.}
  \label{fig:human-eval-2d}
\end{figure}

\begin{figure}[h]
  \centering
  \includegraphics[width=0.85\linewidth]{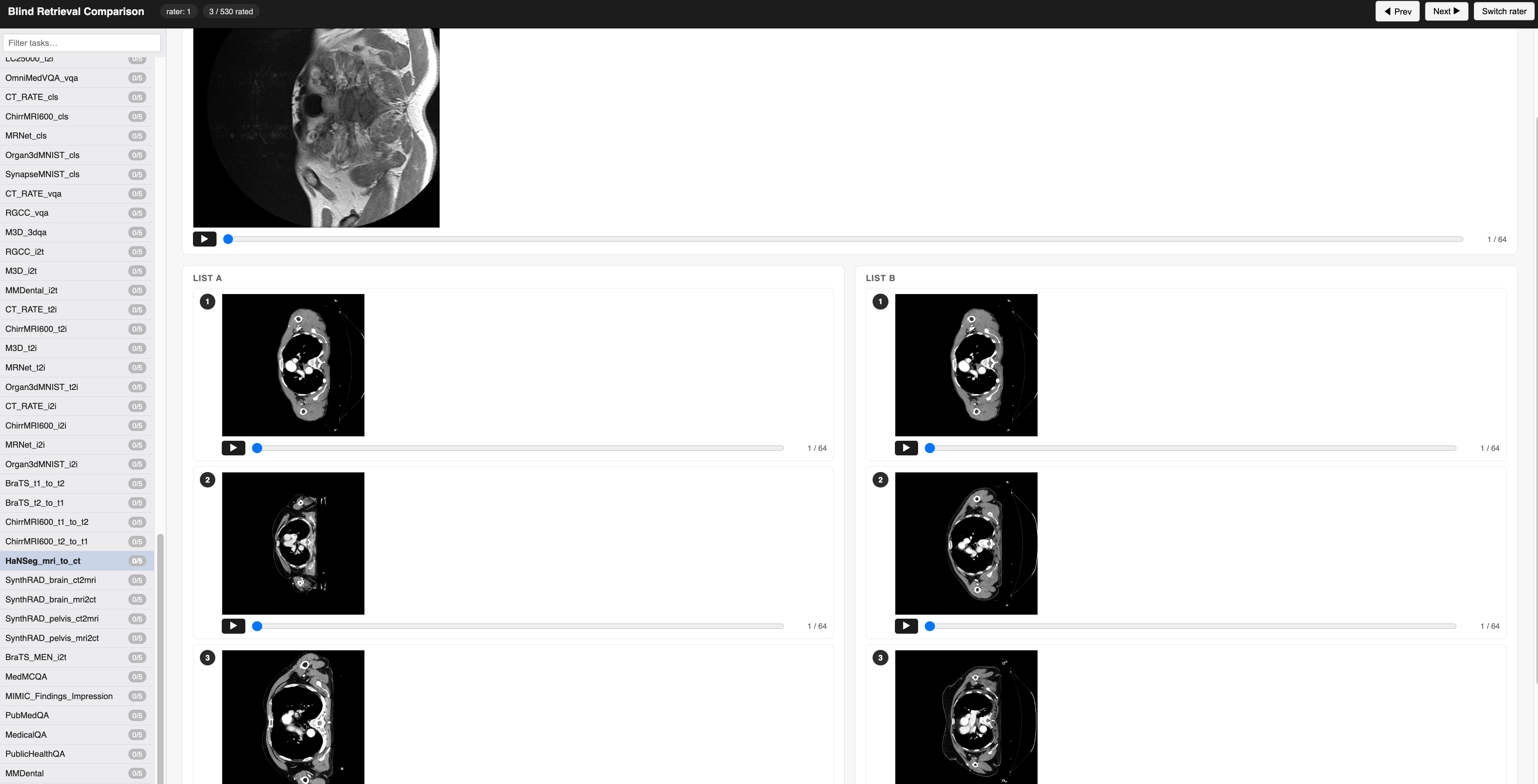}
  \caption{Blind-comparison interface for 3D queries. Volumes are presented as scrubbable slice players; layout, randomization, and preference controls otherwise mirror the 2D interface in Figure~\ref{fig:human-eval-2d}.}
  \label{fig:human-eval-3d}
\end{figure}

\paragraph{Interface.}
The evaluator uses a custom blind-comparison web interface (Figures~\ref{fig:human-eval-2d} and~\ref{fig:human-eval-3d}). The left pane lists all 107 tasks with per-task progress indicators and a filter box; the main pane shows the system prompt, the query (image, 3D volume rendered as a slice player, or text), and the two candidate lists side by side, followed by the three preference buttons and the optional comment field. The same interface is used for 2D and 3D queries; volumes are presented as scrubbable slice players so the evaluator can browse the full volume.

\paragraph{Instructions to the evaluator.}
The evaluator is instructed to select the list that better supports clinical interpretation of the query, considering anatomy, modality, and clinically meaningful similarity. The instructions emphasize that plausibility of returned candidates may matter even when neither list contains the exact gold match, and that ties are permitted when the two lists are judged clinically equivalent.

\paragraph{Evaluator and compensation.}
A single board-certified clinician completed all $530$ comparisons. The clinician participated as a research collaborator on this project; no crowdsourcing platform was used. Compensation arrangements will be disclosed in the final non-anonymous release together with institutional review or exemption status if applicable.

\FloatBarrier
\subsection{Additional Ablations}\label{app:additional-ablations}

\subsubsection{Training Task-Group Ablation}\label{app:task-group-ablation}
Using Qwen3-VL-Embed-2B as the common initialization, we compare mixed-task training with training on only 2D, only 3D, or only text tasks on a fixed randomly sampled subset.
Optimization settings and per-task exposure are held constant, with one epoch per included task; total training exposure varies with the task groups included.
Table~\ref{tab:task-group-ablation} reports this subset-based ablation.
Mixed-task training achieves the highest overall, 2D, and 3D scores, while text-only training is slightly higher on text tasks.

\begin{table}[!htbp]
\centering
\caption{Training task-group ablation on a fixed sampled subset. Scores are Hit@1 (\%).}
\label{tab:task-group-ablation}
\small
\setlength{\tabcolsep}{8pt}
\begin{tabular}{@{}lrrrr@{}}
\toprule
Training configuration & 2D tasks & 3D tasks & Text tasks & Overall \\
\midrule
Pretrained initialization & 21.27 & 16.32 & 58.54 & 22.05 \\
2D tasks only & 32.54 & 16.54 & 56.48 & 29.04 \\
3D tasks only & 24.21 & 17.19 & 57.02 & 24.07 \\
Text tasks only & 20.35 & 15.19 & \textbf{60.92} & 21.27 \\
\rowcolor{MedEmbRow}
Mixed tasks & \textbf{34.28} & \textbf{18.62} & 59.97 & \textbf{30.99} \\
\bottomrule
\end{tabular}
\end{table}

\FloatBarrier
\subsubsection{Instruction-Free Evaluation}\label{app:instruction-free}
We evaluate $49$ tasks whose relevance is defined by the original inputs: $20$ image--image, $22$ image-to-label, and $7$ image-to-report tasks.
For every model, task instructions are removed at evaluation time while the original image and text inputs are retained.
Both MedEmb variants outperform the public baselines in all three families (Table~\ref{tab:instruction-free}).

\begin{table}[!htbp]
\centering
\caption{Instruction-free retrieval on 49 tasks. Scores are task-macro Hit@1 (\%).}
\label{tab:instruction-free}
\small
\setlength{\tabcolsep}{8pt}
\begin{tabular}{@{}lrrrr@{}}
\toprule
Model & I--I (20) & I--Lab (22) & I--Rep (7) & All (49) \\
\midrule
SigLIP & 37.03 & 19.12 & 7.30 & 24.74 \\
BLIP-2 & 33.30 & 10.67 & 5.54 & 19.18 \\
MedImageInsight & 47.50 & 23.57 & 25.10 & 33.55 \\
MedSigLIP & 40.81 & 20.58 & 16.36 & 28.23 \\
RzenEmbed & 31.49 & 13.17 & 7.50 & 19.83 \\
VLM2Vec-V2.0 & 30.97 & 11.40 & 9.22 & 19.07 \\
Qwen3-VL-Embed-2B & 38.15 & 17.41 & 8.41 & 24.59 \\
Qwen3-VL-Embed-8B & 36.50 & 17.89 & 8.28 & 24.11 \\
\midrule
\rowcolor{MedEmbRow}
MedEmb-2B & 52.24 & \textbf{34.82} & 25.76 & 40.64 \\
\rowcolor{MedEmbRow}
MedEmb-8B & \textbf{54.42} & 33.39 & \textbf{28.97} & \textbf{41.34} \\
\bottomrule
\end{tabular}
\end{table}

\FloatBarrier
\subsection{Qualitative Case Studies}\label{app:case-studies}

\begin{figure}[h]
  \centering
  \includegraphics[width=\linewidth]{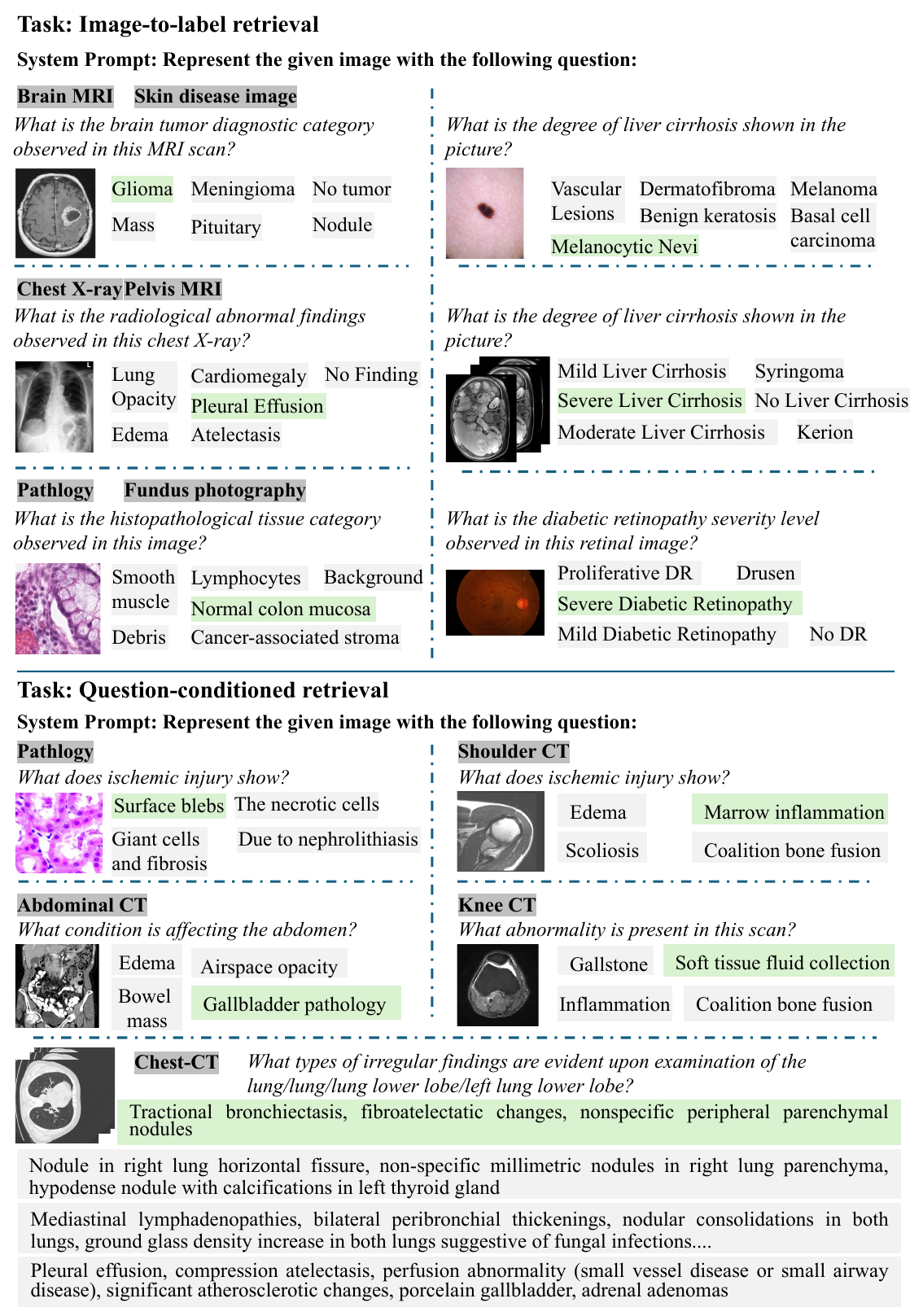}
  \caption{Case study 1.}
  \label{fig:case1}
\end{figure}

\begin{figure}[h]
  \centering
  \includegraphics[width=\linewidth]{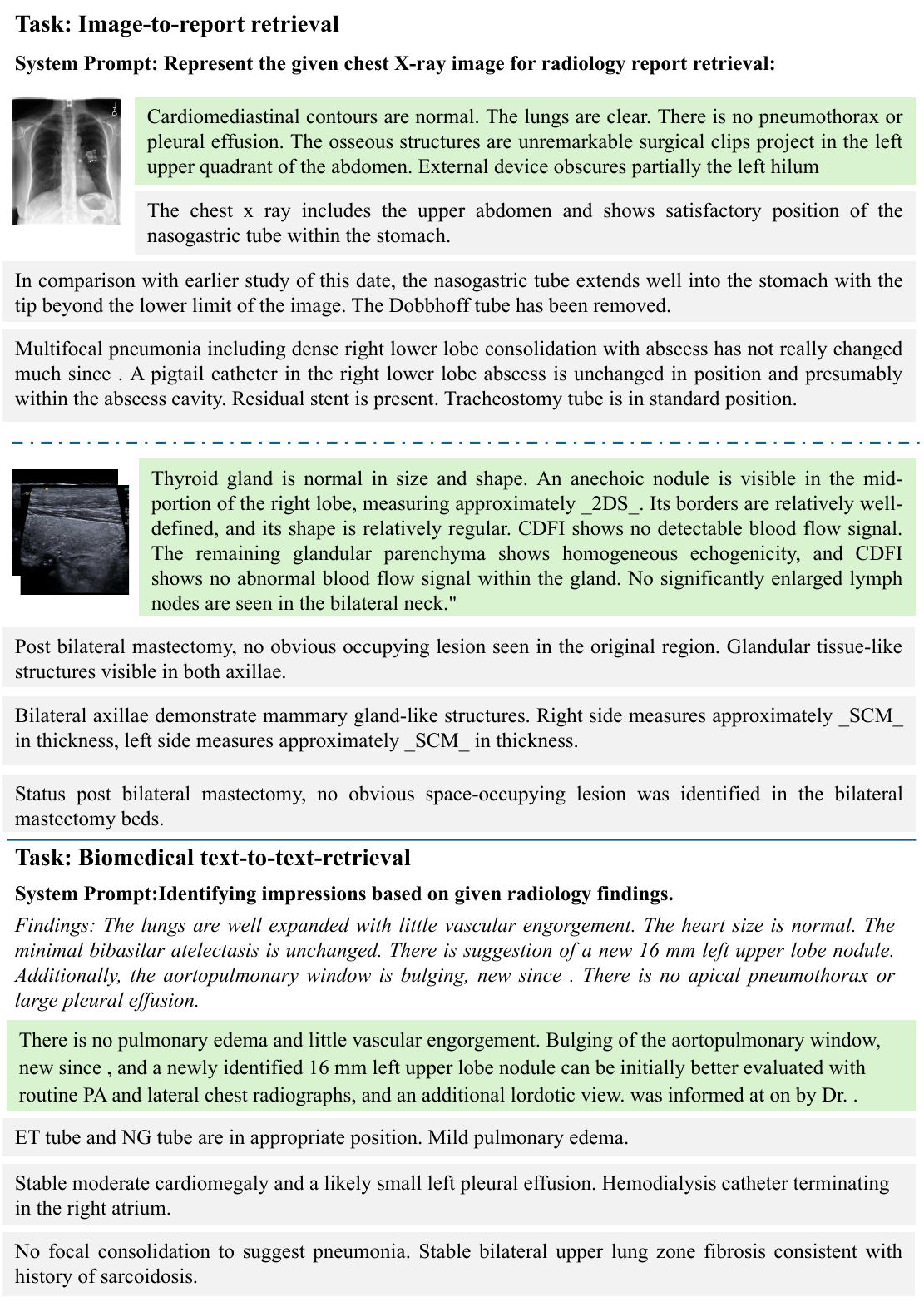}
  \caption{Case study 2.}
  \label{fig:case2}
\end{figure}

\begin{figure}[h]
  \centering
  \includegraphics[width=\linewidth]{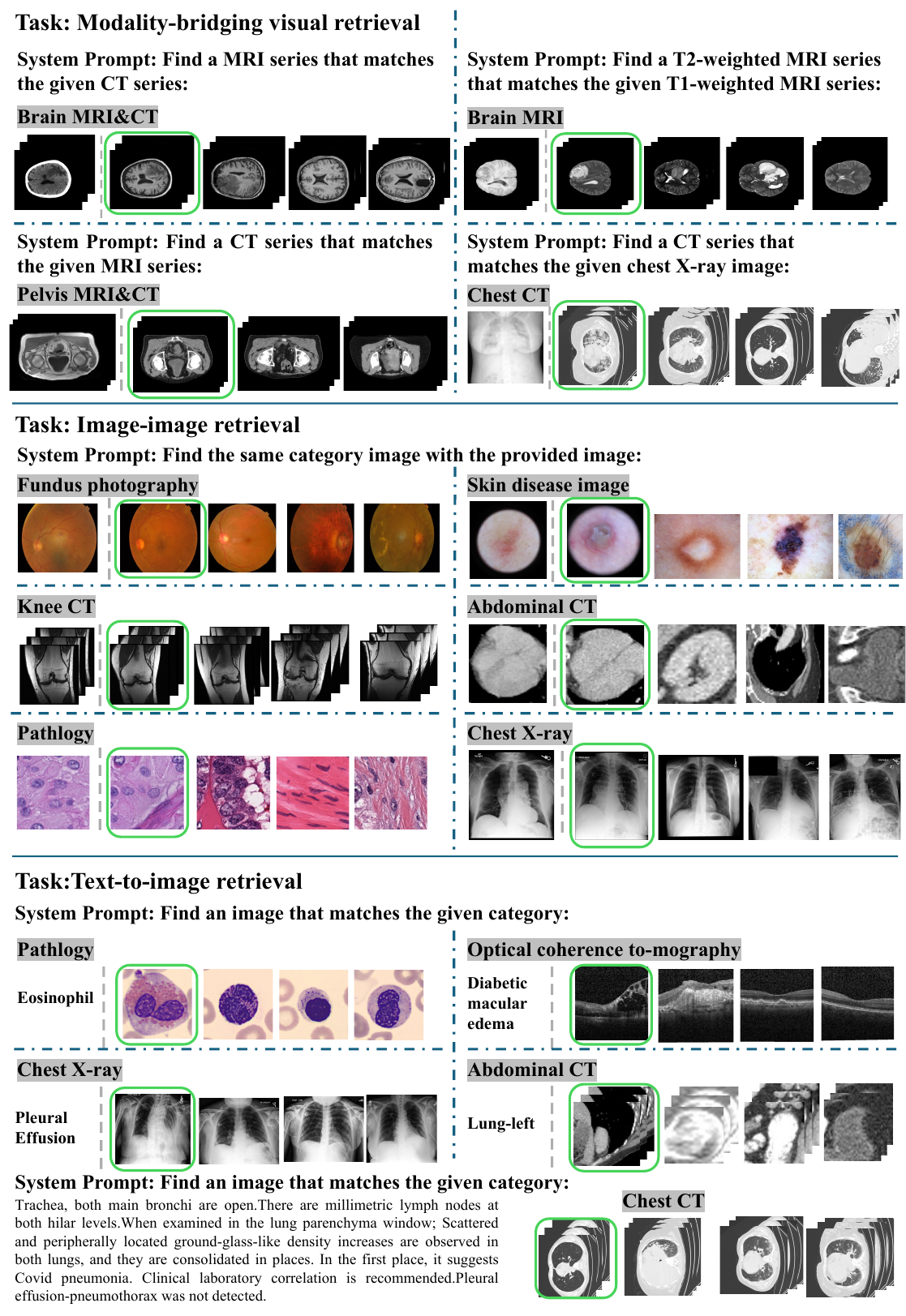}
  \caption{Case study 3.}
  \label{fig:case3}
\end{figure}

\begin{figure}[h]
  \centering
  \includegraphics[width=\linewidth]{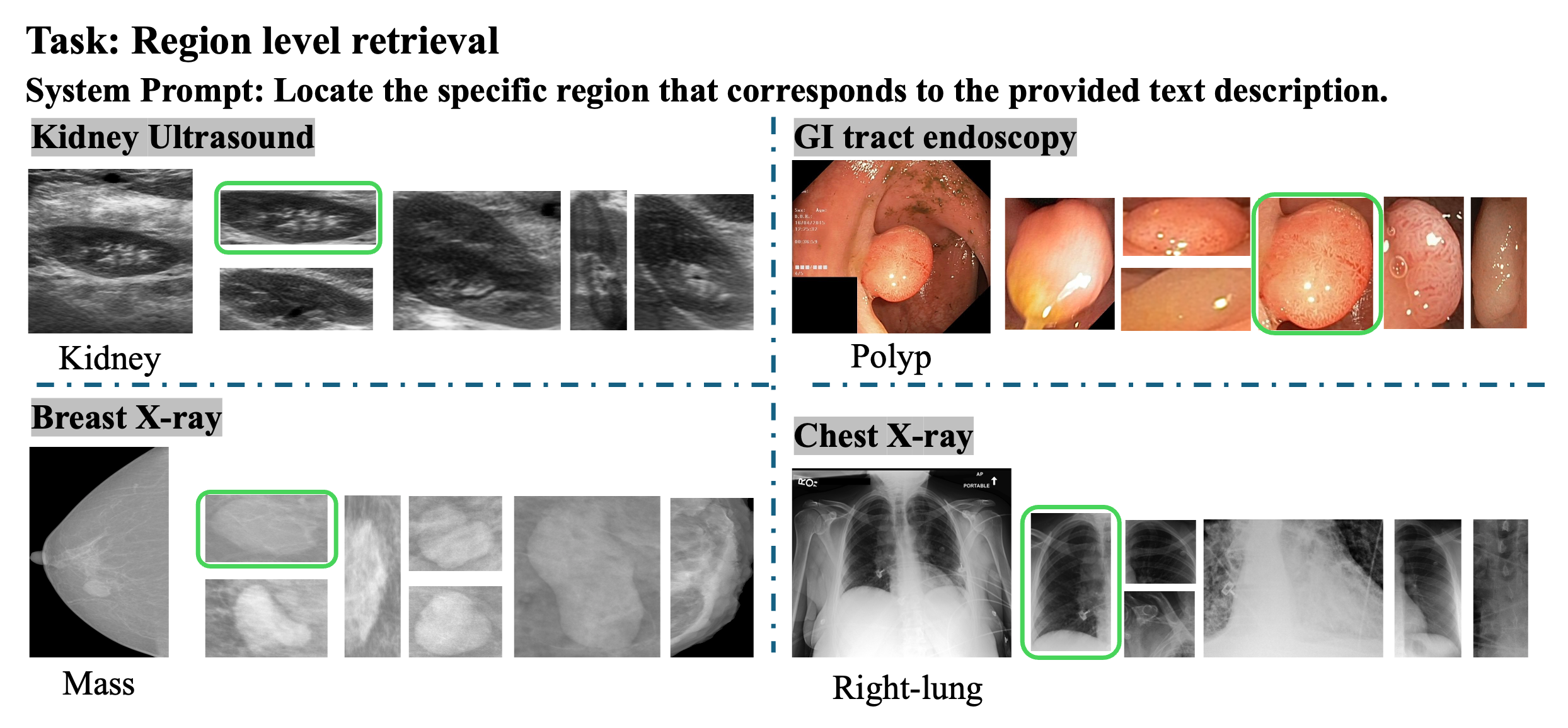}
  \caption{Case study 4.}
  \label{fig:case4}
\end{figure}

Figures~\ref{fig:case1}--\ref{fig:case4} show MedEmb retrieval examples that together cover all eight MedHEB retrieval families. Each block displays the system prompt, the query, and the candidate pool; the gold positive is highlighted in green. The cases span both established families (image--image, image-to-label, image-to-report, text-to-image, biomedical text--text) and underexplored clinical-relation families (modality-bridging visual retrieval, question-conditioned retrieval, region-level text-to-region retrieval), and across modalities including chest X-ray, brain/abdominal/knee/chest CT, brain/pelvis MRI, ultrasound, fundus photography, OCT, dermatology, pathology, and GI endoscopy.

\clearpage
\section*{NeurIPS Paper Checklist}



\begin{enumerate}

\item {\bf Claims}
    \item[] Question: Do the main claims made in the abstract and introduction accurately reflect the paper's contributions and scope?
    \item[] Answer: \answerYes{} 
    \item[] Justification: The abstract and introduction state the benchmark scope, MedEmb reference baseline, and the main empirical conclusion that current medical embeddings are uneven across retrieval families; these claims are substantiated in Sections~\ref{sec:intro}, \ref{sec:main-results}, and \ref{sec:analysis}

\item {\bf Limitations}
    \item[] Question: Does the paper discuss the limitations of the work performed by the authors?
    \item[] Answer: \answerYes{} 
    \item[] Justification: Appendix~\ref{app:release} explicitly discusses dataset noise, modality imbalance, access restrictions, the single-positive ranking formulation, and the scope of MedEmb as a reference baseline rather than a deployment-ready system.

\item {\bf Theory assumptions and proofs}
    \item[] Question: For each theoretical result, does the paper provide the full set of assumptions and a complete (and correct) proof?
    \item[] Answer: \answerNA{} 
    \item[] Justification: The paper is empirical and benchmark-oriented and does not present theorems, formal propositions, or proofs requiring a theory appendix.

    \item {\bf Experimental result reproducibility}
    \item[] Question: Does the paper fully disclose all the information needed to reproduce the main experimental results of the paper to the extent that it affects the main claims and/or conclusions of the paper (regardless of whether the code and data are provided or not)?
    \item[] Answer: \answerYes{} 
    \item[] Justification: The paper specifies task construction, split design, model interfaces, metrics, hyperparameters, and MedEmb training details in Sections~\ref{sec:construction}, \ref{sec:baseline}, \ref{sec:metrics}, and Appendix~\ref{app:medemb-details}, which are the ingredients needed to reproduce the reported benchmark results.

\item {\bf Open access to data and code}
    \item[] Question: Does the paper provide open access to the data and code, with sufficient instructions to faithfully reproduce the main experimental results, as described in supplemental material?
    \item[] Answer: \answerYes{} 
    \item[] Justification: Appendices~\ref{app:release} and \ref{app:datasets} describe the released artifact (task indices, preprocessing scripts, evaluation code, and source-specific access manifests) and list the official access routes for the underlying datasets.

\item {\bf Experimental setting/details}
    \item[] Question: Does the paper specify all the training and test details (e.g., data splits, hyperparameters, how they were chosen, type of optimizer) necessary to understand the results?
    \item[] Answer: \answerYes{} 
    \item[] Justification: Sections~\ref{sec:baseline} and \ref{sec:metrics} report the optimizer, learning-rate schedule, LoRA configuration, training duration, batching strategy, model families, metrics, and evaluation protocol, with additional objective details in Appendix~\ref{app:medemb-details}.

\item {\bf Experiment statistical significance}
    \item[] Question: Does the paper report error bars suitably and correctly defined or other appropriate information about the statistical significance of the experiments?
    \item[] Answer: \answerYes{} 
    \item[] Justification: Main benchmark metrics are computed deterministically over fixed query/candidate pools, and Appendix~\ref{app:release} clarifies this evaluation regime; for the clinician preference study, Section~\ref{sec:human-eval} and Appendix~\ref{app:release} report an exact binomial comparison of non-tied wins ($260/382$, $p<10^{-11}$).

\item {\bf Experiments compute resources}
    \item[] Question: For each experiment, does the paper provide sufficient information on the computer resources (type of compute workers, memory, time of execution) needed to reproduce the experiments?
    \item[] Answer: \answerYes{} 
    \item[] Justification: We report the compute resources used for training the reference model. The benchmark evaluation is lightweight: all MedHEB test experiments can be executed on a single GPU, with peak memory usage below 24GB.
    
\item {\bf Code of ethics}
    \item[] Question: Does the research conducted in the paper conform, in every respect, with the NeurIPS Code of Ethics \url{https://neurips.cc/public/EthicsGuidelines}?
    \item[] Answer: \answerYes{} 
    \item[] Justification: The work is a benchmark and reference-model study over previously released medical datasets, does not claim autonomous clinical deployment, and explicitly discusses release scope, access restrictions, and the need for human oversight in Appendix~\ref{app:release}.

\item {\bf Broader impacts}
    \item[] Question: Does the paper discuss both potential positive societal impacts and negative societal impacts of the work performed?
    \item[] Answer: \answerYes{} 
    \item[] Justification: Appendix~\ref{app:release} discusses positive impacts (more reproducible evaluation and clearer failure analysis) as well as risks such as overinterpreting benchmark scores, propagating source-data bias, and misusing retrieved neighbors as unsupported clinical decision aids.
    
\item {\bf Safeguards}
    \item[] Question: Does the paper describe safeguards that have been put in place for responsible release of data or models that have a high risk for misuse (e.g., pre-trained language models, image generators, or scraped datasets)?
    \item[] Answer: \answerYes{} 
    \item[] Justification: Appendix~\ref{app:release} describes a release scope that excludes restricted raw medical data and instead distributes task indices, code, and source-specific access manifests, so downstream users must still obtain the assets under the original access terms and clinical governance.

\item {\bf Licenses for existing assets}
    \item[] Question: Are the creators or original owners of assets (e.g., code, data, models), used in the paper, properly credited and are the license and terms of use explicitly mentioned and properly respected?
    \item[] Answer: \answerYes{} 
    \item[] Justification: The paper credits source datasets and official access URLs throughout the paper and Appendix~\ref{app:datasets}, but it does not enumerate every underlying license or term-of-use in the manuscript itself; those asset-level license details should accompany the final artifact release.

\item {\bf New assets}
    \item[] Question: Are new assets introduced in the paper well documented and is the documentation provided alongside the assets?
    \item[] Answer: \answerYes{} 
    \item[] Justification: The paper documents the new benchmark, task construction protocol, release scope, and MedEmb training recipe in Sections~\ref{sec:construction}--\ref{sec:main-results} and Appendices~\ref{app:medemb-details} and \ref{app:release}.

\item {\bf Crowdsourcing and research with human subjects}
    \item[] Question: For crowdsourcing experiments and research with human subjects, does the paper include the full text of instructions given to participants and screenshots, if applicable, as well as details about compensation (if any)?
    \item[] Answer: \answerYes{}
    \item[] Justification: The clinician preference study (Section~\ref{sec:human-eval}) is documented in Appendix~\ref{app:human-protocol}, which describes the sampling procedure (530 queries stratified across 107 tasks), per-query A/B randomization, the full text of the rater instructions, and screenshots of the blind-comparison interface for 2D and 3D queries (Figures~\ref{fig:human-eval-2d} and~\ref{fig:human-eval-3d}). No crowdsourcing platform was used; the rater is a single board-certified clinician collaborator. Compensation arrangements will be disclosed in the final non-anonymous release.

\item {\bf Institutional review board (IRB) approvals or equivalent for research with human subjects}
    \item[] Question: Does the paper describe potential risks incurred by study participants, whether such risks were disclosed to the subjects, and whether Institutional Review Board (IRB) approvals (or an equivalent approval/review based on the requirements of your country or institution) were obtained?
    \item[] Answer: \answerNA{} 
    \item[] Justification: The clinician preference study involved a single expert collaborator evaluating model outputs without collection of personal, behavioral, or sensitive participant data. No patient interaction or identifiable patient information was involved.

\item {\bf Declaration of LLM usage}
    \item[] Question: Does the paper describe the usage of LLMs if it is an important, original, or non-standard component of the core methods in this research? Note that if the LLM is used only for writing, editing, or formatting purposes and does \emph{not} impact the core methodology, scientific rigor, or originality of the research, declaration is not required.
    \item[] Answer: \answerNA{} 
    \item[] Justification: LLMs were used only for routine writing assistance, including grammar checking and language consistency checks. Their use did not affect the core method development, experimental design, data analysis, results, or scientific claims of this work.

\end{enumerate}

\end{document}